\documentclass{article}
\usepackage[utf8]{inputenc}

\usepackage[pdfencoding=auto]{hyperref}

\usepackage{geometry}
\usepackage{graphicx}

\usepackage{natbib}
\usepackage{breakcites}
\usepackage{algorithm}
\usepackage{etoolbox}

\usepackage{amsmath}
\usepackage{amssymb}
\usepackage{nicefrac}
\usepackage{float}
\usepackage{hyperref}

\usepackage{xcolor} 

\usepackage{thmtools} 

\declaretheorem[name=Hypothesis, preheadhook={}]{hyp} 

\usepackage{cleveref} 
\crefname{hyp}{hypothesis}{hypotheses} 
\Crefname{hyp}{Hypothesis}{Hypotheses} 

\usepackage{ragged2e} 

\usepackage{booktabs} 
\usepackage{multirow} 
\usepackage{pdflscape} 
\usepackage{algorithm}
\usepackage{algpseudocode}

\algnewcommand{\IfThen}[2]{
	\State \algorithmicif\ #1\ \algorithmicthen\ #2}
\algnewcommand{\LineComment}[1]{\State \(\triangleright\) #1}

\algnewcommand{\InlineFor}[2]{
	\State \algorithmicfor\ #1\ \algorithmicendfor\ #2}

 \makeatletter
\AfterEndEnvironment{algorithm}{\let\@algcomment\relax}
\AtEndEnvironment{algorithm}{\kern2pt\hrule\relax\vskip3pt\@algcomment}
\let\@algcomment\relax
\newcommand\algcomment[1]{\def\@algcomment{\footnotesize#1}}

\renewcommand\fs@ruled{\def\@fs@cfont{\bfseries}\let\@fs@capt\floatc@ruled
  \def\@fs@pre{\hrule height.8pt depth0pt \kern2pt}%
  \def\@fs@post{}%
  \def\@fs@mid{\kern2pt\hrule\kern2pt}%
  \let\@fs@iftopcapt\iftrue}
\makeatother

\usepackage[utf8]{inputenc}
\usepackage[T1]{fontenc}

\usepackage{booktabs}

\usepackage[parfill]{parskip} 

\usepackage{authblk}

\usepackage{algorithm}
\usepackage{algpseudocode}

\providecommand{\keywords}[1]
{
  \small	
  \textbf{\textit{Keywords---}} #1
}

\title{Feature-Guided Metaheuristic with Diversity Management for Solving the Capacitated Vehicle Routing Problem}

\author[1]{Bachtiar Herdianto\footnote{Corresponding Author}\footnote{Alternative email: \href{mailto:bachtiarherdianto@gmail.com}{bachtiarherdianto@gmail.com}}}
\author[1]{Romain Billot}
\author[2]{Flavien Lucas}
\author[3]{Marc Sevaux}

\date{\color{blue} 
      \nolinkurl{bachtiar.herdianto@imt-atlantique.fr}$^{*\dagger1}$\\
      \nolinkurl{romain.billot@imt-atlantique.fr}$^1$\\
      \nolinkurl{flavien.lucas@imt-nord-europe.fr}$^2$\\
      \nolinkurl{marc.sevaux@univ-ubs.fr}$^3$}

\affil[1]{IMT Atlantique, Lab-STICC (UMR 6285, CNRS), Brest, France}
\affil[2]{IMT Nord Europe, CERI Systèmes Numériques, Douai, France}
\affil[3]{Université Bretagne Sud, Lab-STICC (UMR 6285, CNRS), Lorient, France}

\begin{document}

\maketitle

\begin{abstract}
We propose a feature-based guidance mechanism to enhance metaheuristic algorithms for solving the Capacitated Vehicle Routing Problem (CVRP). This mechanism leverages an Explainable AI (XAI) model to identify features that correlate with high-quality solutions. These insights are used to guide the search process by promoting solution diversity and avoiding premature convergence. The guidance mechanism is first integrated into a custom metaheuristic algorithm, which combines neighborhood search with a novel hybrid of the split algorithm and path relinking. Experiments on benchmark instances with up to $30,000$ customer nodes demonstrate that the guidance significantly improves the performance of this baseline algorithm.

Furthermore, we validate the generalizability of the guidance approach by integrating it into a state-of-the-art metaheuristic, where it again yields statistically significant performance gains. These results confirm that the proposed mechanism is both scalable and transferable across algorithmic frameworks.
\end{abstract}

\keywords{Metaheuristic Algorithm, Supervised Machine Learning, Explainable Artificial Intelligence (XAI), Capacitated Vehicle Routing Problem}

\section{Introduction}
\label{sec:introduction}
    Routing represents a significant activity in logistics and supply chains, involving the movement of products or goods, from one location to another. This process is crucial for driving economic and social activities, impacting various aspects of daily lives \citep{arnold2019makes}. The main challenge arises from the increasing delivery costs, that may directly influence the pricing of goods. Optimizing the delivery routes becomes one significant solution for mitigating this problem \citep{Simchi-Levi2000-fd}. Various routing problems exist, with one of the most studied being the Capacitated Vehicle Routing Problem (CVRP) \citep{toth2014vehicle, sorensen2013astatistical, Prodhon2016, arnold2019knowledge, accorsi2021fast}. Yet, despite decades, the CVRP remains a challenging problem \citep{prins2004simple,laporte2009fifty}. Recently, there has been a growing interest in using Machine Learning (ML) to enhance optimization algorithms \citep{vinyals2015pointer,Hottung_Andr__2020,bengio2021machine}. However, many optimization solvers start from scratch for similar problem types, ignoring insights from previous solutions. Hence, leveraging past solutions could offer efficient and effective ways to improve optimization algorithms \citep{arnold2019makes}. Thus, the optimization solver can learn from its own decisions, adapting its behavior to refine the performance. In parallel, Explainable Artificial intelligence (XAI) offers techniques that can identify the main features, as well as investigate they behave for making a decision (Lundberg et al., \citeyear{lundberg2017unified,lundberg2020local}, Arrieta et al., \citeyear{arrieta2020explainable}). 

    In this research, we aim to solve the CVRP by developing a hybrid ML and metaheuristic algorithm. Our approach involves developing a learning model that can learn how to achieve an optimal quality solution, based on the problem features. Subsequently, we try to interpret the developed learning model and use these insights to formulate a guidance able to boost the performance of the metaheuristic algorithm.

\subsection{Problem Description and Related Work} 
\label{subsec:relatedwork}
    The CVRP can be defined on an undirected graph $G=(V,E)$, where $V$ includes a depot $D$ and a set of customers $C = \{c_1, c_2, \dots, c_N\}$, with $N=|V|-1$. Each customer has a demand, and weighted edges $E$ represent distances between nodes. The neighborhood of customer $c_i$ is denoted as $\mathcal{N}(c_i)$ \citep{Prodhon2016}. A solution is a set of routes starting and ending at the depot, serving customers without exceeding vehicle capacity $Q$. It is feasible if each customer is visited exactly once. The goal is to minimize total route cost \citep{laporte2009fifty}. The CVRP is a basic form of the VRP and can be extended to more complex variants, such as the VRP with Time Windows (VRPTW) \citep{solomon1987algorithms}, where $c_i$ must be visited within a specified time interval.

    \paragraph{\textbf{Metaheuristics for solving the CVRP}} 
        Heuristics and local search are key components of metaheuristics \citep{Prodhon2016}. Tabu Search \citep{glover1997tabu}, helps escape local optima by avoiding cycles. Granular Neighborhoods (GNs) \citep{toth2003granular}, filter out less promising neighbors and, when combined with Tabu Search, offer a strong balance between speed and quality, even on problems with up to $30,000$ customers \citep{accorsi2021fast}. To further enhance intensification while maintaining diversity, path relinking \citep{glover1997tabu,glover2000scatter} is often integrated. In parallel, \cite{prins2004simple} introduced a new representation of the solution, called the giant tour, and a splitting mechanism to transform it into a VRP solution, and was later refined for many VRP variants \citep{VIDAL2014658}, including CVRP instances with up to $1,000$ customers \citep{vidal2022hybrid}. While path relinking has been shown to strengthen Tabu Search \citep{laguna1999intensification,ho2006path}, its added value appears limited when hybridized with GRASP and VND, compared to simpler GRASP–VND combinations \citep{sorensen2013astatistical}.

    \paragraph{\textbf{Learning algorithm for optimization}} 
        The integration of machine learning (ML) with optimization can follow three strategies \citep{bengio2021machine}: (1) end-to-end learning, (2) learning from problem properties, and (3) learning repeated decisions. The second strategy uses ML to configure the optimization algorithm, while the third embeds ML into the algorithm loop to adapt behavior from prior decisions. End-to-End learning includes neural-based VRP solvers \citep{ma2023learning}, such as Learning-to-Construct (L2C), Learning-to-Search (L2S), and Learning-to-Predict (L2P). L2C builds solutions step-by-step using models like Pointer Networks \citep{vinyals2015pointer} or attention mechanisms \citep{kool2018attention}, and can be accelerated with methods like Efficient Active Search (EAS) \citep{hottung2022efficient}. L2S \citep{wu2021learning} refines solutions via search heuristics (\textit{e.g.}, k-opt) but is computationally demanding. L2P \citep{joshi2019efficient,kool2022deep} predicts critical patterns (\textit{e.g.}, edge heatmaps), scaling well to larger instances, though limited by supervised learning and difficulty handling complex constraints.
    
    \paragraph{\textbf{Hybridizing machine learning with metaheuristic}}
        The hybrid mechanism generally refers to using machine learning (ML) approach to enhance optimization algorithms. ML can initially be applied to analyze problem structure and then use the extracted insights to enhance the optimization process \citep{zhang2022deep,parmentier2023structured,zarate2025machine}. 
        
        This form of learning to configure algorithms \citep{bengio2021machine} combines data-driven analysis with metaheuristic design. Models to distinguish near-optimal from non-optimal solutions using both instance-level and solution-based features to guide the search process have been proposed in prior studies \citep{arnold2019makes, lucas2019comment}. Later, this knowledge was applied to improve CVRP solvers, including for large-scale problem variants \citep{arnold2019knowledge,ARNOLD201932}. However, previous works did not provide a statistical analysis to isolate the impact of the guidance mechanism on overall algorithm performance. Despite this limitation, both research found that solution-based features were generally more predictive than instance features \citep{arnold2019makes,lucas2019comment}. This is further supported that ranking the relative importance of problem attributes can be used to define effective guidance rules \citep{guidotti2018survey}. These findings point toward future opportunities to enhance metaheuristics through machine learning–driven guidance mechanisms.

    \subsection{Research Questions and Contributions}
    \label{subsec:research-question-and-contributions}
        \cite{arnold2019knowledge} has shown that clearly defining the preferred structural properties of an optimal VRP solution is highly valuable for designing an efficient heuristic. Furthermore, with more interpretation by using an explainable learning model, we can change our way of solving VRP \citep{lucas2019comment}. An ML model, through the learning phase, builds predictive models that can map data features into a class \citep{guidotti2018survey}. Then, the explanation of these models gives insights into how the models utilize features to make decisions. Meanwhile, the ranking of the relative importance of the problem attributes can be incorporated effectively \citep{guidotti2018survey}. Inspired by those advances, in this research, our main questions are:
        
        \begin{enumerate}
            \item How can we extract the most important features of a good solution and use them to guide a heuristic?
            \item To what extent does the developed heuristic bring a significant improvement for solving the CVRP?
        \end{enumerate}

        To address these questions, we propose a simple mechanism of hybridization between ML and metaheuristics by following the design of \textit{learning to configure algorithms}. We introduce an explainable learning framework for classifying the quality of the VRP solution. This is done by generating a dataset of VRP features and developing a classification model that can identify the features that have the most significant influence, then explaining the developed model to study the feature that influences the quality of the solution. Furthermore, a metaheuristic for solving the CVRP that present a novel mechanism of path relinking that hybridizes with the split algorithm is also introduced. Finally, we present a feature-based guidance applied to the proposed metaheuristic. In summary, the main design steps and contributions of this research are the following: 

        \begin{enumerate}
            \item We generate a dataset of features from $10,000$ $\mathbb{XML}$100 VRP instances, including optimal and near-optimal solutions \citep{queiroga202110}.
            \item We develop an ML model to classify solution quality and identify key distinguishing features.
            \item For leveraging these knowledge, we formulate a guidance for managing solution diversity in a metaheuristic algorithm.
            \item Moreover, a new metaheuristic algorithm for solving the CVRP is introduced as a baseline for leveraging our proposed guidance. This metaheuristic presents a novel path relinking mechanism, that integrates giant tour concatenation with a hybridized split algorithm.
            \item The computational experiment show that the proposed guidance mechanism able to improve the performance the baseline metaheuristic. 
            \item Lastly, we generalize the proposed guidance mechanism to enhance the performance of across different baseline for demonstrating its adaptability.
        \end{enumerate}

        The rest of this paper is structured as follows: \Cref{sec:learning-from-solution} outlines the explainable learning framework for classifying quality of solution for developing the proposed guidance. In \Cref{sec:development-metaheuristic} describe our proposed metaheuristic as the initial baseline for our proposed guidance. \Cref{sec:hybridizing-ml-mh} details its hybridization of our metaheuristic algorithm with the proposed guidance. Lastly, \cref{sec:experiment-analysis} shows our experimentation to evaluate the proposed feature-based guidance.

\section{Learning From Solutions} 
\label{sec:learning-from-solution}
    \citet{arnold2019knowledge} used Support Vector Machine (SVM) to predict solution quality in the VRP. They defined 18 features, both instance and solution, and argued that solution features are generally more important than instance features. Similarly, \citet{lucas2019comment} demonstrated that not all features contribute meaningfully to prediction accuracy. 

\begin{figure}[htbp]
    \begin{center}
        \includegraphics[width=0.98\textwidth]{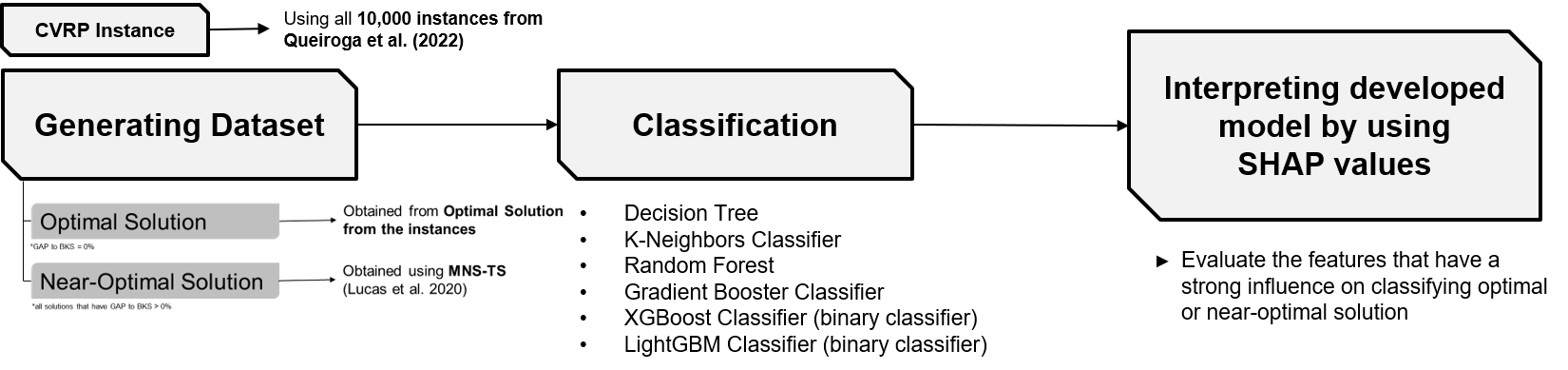}
        \caption{Outline of the learning framework.}
        \label{img:pipeline}
    \end{center}
\end{figure}
    
    Moreover, \citet{arnold2019knowledge} achieved strong predictive performance, particularly on larger instances with more diverse solutions. They identified compactness, route width and angular span, the number of intersecting edges, and the distances of connecting edges to the depot as key distinguishing features between near-optimal and non-optimal solutions. In line with this, \citet{lucas2019comment} used statistical techniques such as PCA to determine which features were most relevant to solution quality. Their findings supported the conclusion from \citet{arnold2019knowledge}. Furthermore, \citet{lucas2019comment} emphasized the potential of leveraging feature importance and correlations as a foundation for developing guidance for metaheuristic. Building on these insights, this section explores the use of Explainable AI (XAI) techniques to extend the study from \citet{arnold2019knowledge} and \citet{lucas2019comment}. Thus, our proposed methodology involves three main steps: (1) generating a dataset, (2) performing classification using various classifier models, and (3) explaining the resulting models using SHAP values, as illustrated in \Cref{img:pipeline}. The dataset alongside the source code and the documentation for performing the analysis are available\footnote{\url{https://github.com/bachtiarherdianto/MS-Feature}}. Let $\mathcal{X}$ be the set of $N$ training samples, where each sample has $p$ features and a corresponding label. Then, $x^{(i)}$ denote the feature vector of the $i$-th sample, with $x^{(i)} = [x_1^{(i)}, \dots, x_p^{(i)}]$. The corresponding label $y^{(i)} \in \{0, 1\}$, for $i = 1, 2, \dots, N$, where $y^{(i)} = 1$ indicates that the solution is categorized as \textit{optimal}, and $y^{(i)} = 0$ indicates it is categorized as \textit{near-optimal}. The goal is to learn a model $f(x)$ that predicts the label $y$ from the features $x$, such that:
\begin{equation} \label{eq:f-x}
    f(x) = 
    \begin{cases} 
        1 & \text{{\small if it corresponds to an optimal solution}}\\ 
        0 & \text{{\small if it corresponds to a near-optimal solution}} 
    \end{cases}
\end{equation}

\begin{figure}[htbp]
    \begin{center}
        \includegraphics[width=0.55\textwidth]{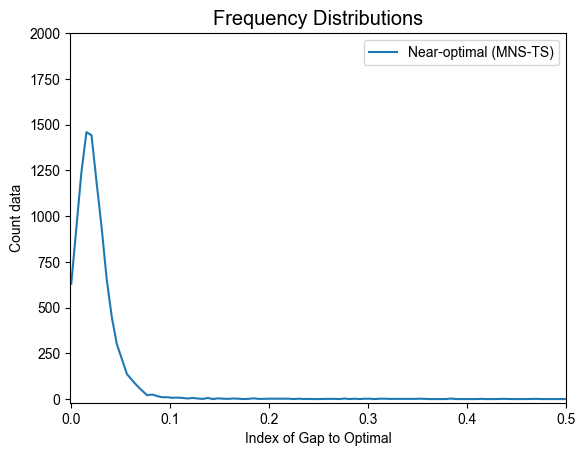}
        \caption{The summary of solutions from the MNS-TS algorithm with a 10-second time budget, showing mostly had a gap to the optimal solution between $0.0$ and $0.5$.}
        \label{img:input-data}
    \end{center}
\end{figure}

    \subsection{Data Generations and Feature Extractions}
    \label{subsec:data-generation} 
        The CVRP has a great variety of instances according to the following attributes: (1) the positioning and number of customers, (2) the positioning of the depot, (3) the distribution of demand, and (4) the average route size or the number of routes \citep{UCHOA2017845}. In this research, we use the $10,000$ $\mathbb{XML}$100 instances\footnote{Detailed information related to the problem instances is available at \url{https://galgos.inf.puc-rio.br/cvrplib/en/instances}} \citep{queiroga202110} to generate a dataset that is used to develop a learning model $f(x)$. To generate near-optimal solutions, we utilize the MNS-TS algorithm \citep{soto2017multiple,lucas2020reducing}, a Multiple Neighborhood Search combined with Tabu Search, previously shown effective for the OVRP \citep{soto2017multiple}. Each instance is solved within a $T_{max}=10$ second time limit, repeated over five runs. The performance is assessed using the relative difference between the MNS-TS solution and the known optimal objective value, calculated as follows: 
\begin{equation} \label{eq:gap-to-opt}
    \text{Gap-to-Optimal} := \dfrac{\text{Obtained Solution} - \text{Optimal Solution}}{\text{Optimal Solution}} \times 100\%
\end{equation}
        The dataset used for developing the learning model consists of $20,000$ data points, where $50\%$ of them are optimal solutions, and the remaining are near-optimal solutions.

        \paragraph{\textbf{Feature extractions}} 
             For developing the learning model, features are grouped into: (1) instance features and (2) solution features. Most instance and solution features are based on those proposed in previous studies \citep{arnold2019makes, lucas2019comment, lucas2020reducing}. Features are prefixed with $\text{I}$ for instance-related and $\text{S}$ for solution-related ones. In total, $9$ instance features and 22 solution features are used, detailed in \ref{sec:appendix-feature-vrp}.
        
    \subsection{Learning Model: Binary Classification}
    \label{subsec:learning-model}
        To guide the metaheuristic process, we trained several supervised classification models to differentiate between high-quality and suboptimal CVRP solutions. The candidate algorithms included $K$-Nearest Neighbors \citep{cover1967nearest}, Decision Tree Classifier \citep{breiman2017classification}, and Random Forest \citep{breiman2001random}, as well as boosting-based methods such as Gradient Boosting \citep{friedman2001greedy}, Extreme Gradient Boosting (XGBoost) \citep{chen2016xgboost}, and LightGBM \citep{ke2017lightgbm}. 
        To ensure valid model training and prevent overfitting, the dataset was partitioned by instance: $70\%$ of instances were allocated for training and feature selection, while $30\%$ were retained for out-of-sample testing. We further applied instance-based $k$-fold cross-validation to evaluate generalization. This approach avoids data leakage and overfitting. 
        
        All models were trained using our VRP feature dataset, implemented in Python, and executed on a $64$-bit machine equipped with an AMD Ryzen 7 PRO 5850U processor and $16$ GB of RAM, running Ubuntu 22.04.1. Model performance was evaluated using the $F_1$-score, calculated as follows:
\begin{equation} \label{eq:f1-score}
    F_1\text{-score} := \dfrac{2 \cdot \text{precision} \cdot \text{recall}}{\text{precision} + \text{recall}}
\end{equation} 
        
        A detailed comparison, including precision and recall, is presented in \Cref{table:f1score}. Among the tested models, the Gradient Boosting classifier achieved the highest $F_1$-score on the held-out test set. Although the accuracy remains modest, predictive accuracy is not the primary objective. Rather, our aim is to extract predictive features that reflect the structural quality of solutions. The following section investigates how the Gradient Boosting model forms predictions, focusing on feature contributions and interpretability.

\begin{table}[htbp]
\caption{$F_1$-scores from various classification algorithms used in this proposed model.}
\label{table:f1score}
\vspace*{0.1cm}
\centering
\scalebox{0.76}
{
    \begin{tabular}{lcccc}
        \toprule
            \textbf{Algorithm}                      & \textbf{Precision}  & \textbf{Recall}   && \textbf{$F_1$-score}\\
        \midrule
            $K$-Nearest Neighbors Classifier          & 0.476               & 0.351             && 0.404\\
            Decision Tree Classifier                & 0.537               & 0.538             && 0.538\\
            Random Forest Classifier                & 0.523               & 0.500             && 0.511\\
            \textbf{Gradient Boosting Classifier}   & \textbf{0.672}      & \textbf{0.661}    && \textbf{0.666}\\
            XGBoost Classifier                      & 0.632               & 0.621             && 0.627\\
            LightGBM Classifier                     & 0.652               & 0.652             && 0.652\\
        \bottomrule
    \end{tabular}
}
\end{table}

    \subsection{Explaining The Learning Model}
    \label{subsec:xplainingp-learning-model}
\begin{figure}[H]
    \begin{center}
        \includegraphics[width=0.94\textwidth]{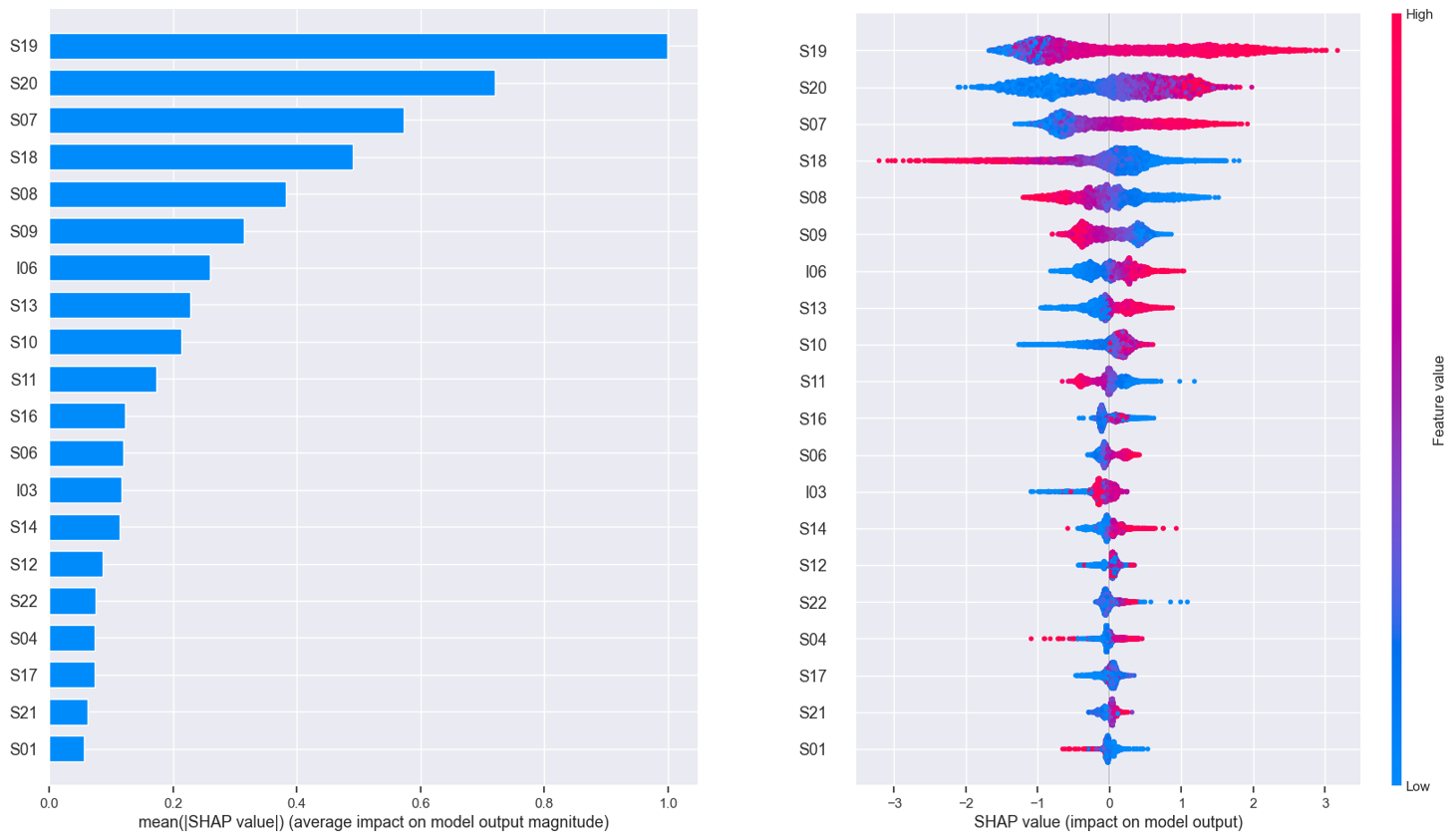}
        \caption{The global feature importance plot (left) and the local explanation summary plot (right) from the learning model.}
        \label{img:beeswarm}
    \end{center}
\end{figure}

        Explainable AI provides insights into how AI models learn and make decisions \citep{arrieta2020explainable}. Among various explainability methods, SHAP (SHapley Additive exPlanations) \citep{lundberg2017unified,lundberg2020local,baptista2022relation} is a notable approach for interpreting model predictions. Based on \Cref{table:f1score}, we compute SHAP values for the Gradient Boosting classifier, with their distribution illustrated in \Cref{img:beeswarm}.

        As shown in the left plot of \Cref{img:beeswarm}, features S19 (average capacity utilization) and S20 (standard deviation of capacity utilization) exhibit the strongest influence among all features, indicates that these two features are important in determining the prediction. The right plot displays SHAP values, quantifying each feature’s impact on the prediction. A higher SHAP value indicates a stronger contribution, while the color gradient represents the magnitude of the feature values: red indicates higher values, while blue corresponds to lower values. 
        
        For S19, high values correlate with positive SHAP values, suggesting that increased capacity utilization generally improves the quality of the solution. For S20, although high values can contribute positively, low values are more frequently associated with negative SHAP impacts. This implies that a high variability in capacity utilization has a more detrimental effect on solution quality than the benefit derived from lower standard deviations. From these observations, we can conclude that maximizing the average capacity utilization (S19) across routes in the obtained solution leads to a higher-quality solution, while minimizing variation in capacity utilization (S20) is beneficial to avoiding a negative impact on the performance of the algorithm. 

    \subsection{Integration Into the Metaheuristic}
    \label{subsec:integrations-mhml}
        Feature importance can be used to design decision rules \citep{guidotti2018survey}. As shown in \Cref{img:beeswarm}, features S19 and S20 exhibit the highest SHAP values, indicating their dominant influence. Based on discussion in \Cref{subsec:xplainingp-learning-model}, we formulate a hypothesis that solution diversity in the elite pool can be guided using these key features when solving the CVRP.
        
        \begin{justify}
            \textbf{Hypothesis: }\textit{features related to capacity utilization can be used to control the diversity of a pool of elite set of solutions}
        \end{justify} 
        
        To test this hypothesis, we develop a metaheuristic baseline using the learning-derived guidance and conduct computational experiments on the $\mathbb{XML}$100 instances in \Cref{subsec:xml-result} to evaluate its impact.

\section{Development of the Baseline Metaheuristic}
\label{sec:development-metaheuristic}
    This section presents a new metaheuristic algorithm for solving the CVRP, named the Multiple Search (MS) algorithm. It extends the MNS-TS method \citep{soto2017multiple,lucas2019comment} and serves as the initial baseline for the proposed guidance mechanism. An overview of the algorithm is presented in \Cref{algorithm:ms-overview}. The MS algorithm consists of a construction phase (line 2) followed by neighborhood search and path relinking. The neighborhood search involves route perturbation and local search to refine solutions. To enhance both diversity and intensification, a hybrid split and path relinking mechanism is introduced. Each component is described in the following sections.
    \begin{algorithm}
        \footnotesize
        \caption{MS for solving the CVRP.}\label{algorithm:ms-overview}
        \hspace*{\algorithmicindent} \textbf{input:} CVRP instance $\mathbf{I}$
        \begin{algorithmic}[1]
            \Procedure{MS-CVRP}{$\mathbf{I}$}
            \State $\mathbb{E} \gets \Call{GeneratingInitialSolutions}{\mathbf{I}}$ \Comment{pool of elite solutions}
            \State $S_{best} \gets \arg \min_{s \in \mathbb{E}} \Call{Cost}{s}$
            \State $\vartheta \gets 0$ \Comment{non-improving iteration counter}
            \Repeat
                \State $(\mathbb{E},S_{best}) \gets \Call{NeighborhoodSearch}{\mathbb{E},S_{best}}$
                \State $(\mathbb{E},S_{best}) \gets \Call{Split-PathRelinking}{\mathbb{E},S_{best}}$
                \State $(\mathbb{E},S_{best}) \gets \Call{EliteSetManagement}{\mathbb{E},S_{best},S_0,\vartheta,\mathbf{I}}$
            \Until{$T_{max}$}
            \State \Return $S_{best}$
            \EndProcedure
        \end{algorithmic}
    \end{algorithm}

    \subsection{Generating Initial Solutions}
    \label{subsec:generating-inisol}
        The initial solution is built using the savings algorithm \citep{clarke1964scheduling}, which can be accelerated by limiting each customer $i \in C$ to $n_{cw}$ nearest neighbors $j \in \mathcal{N}_{cw}(C)$ when computing savings \citep{arnold2019knowledge,accorsi2021fast}. We set $\mathcal{N}_{cw} = 100$ as shown in \Cref{table:paramater-setting}. The algorithm generates at least $\mathbb{E}_{min}$ initial solutions. After constructing a base solution, it applies tour perturbation by destroying two random routes and reinserting their customers into the best positions. 
        
    \subsection{Pool of the Elite Set Solutions} 
    \label{subsec:eliteset}
        The elite set is a pool of high-quality solutions found during the search, with size between $\mathbb{E}_{min}$ and $\mathbb{E}_{max}$ (\Cref{table:paramater-setting}). Initially empty, the pool adds distinct solutions until full. Once full, a better candidate replaces the worst member.

        \subsubsection{Diversity control mechanism} 
        \label{subsubsec:diversity-control}
            The diversity control mechanism controls the level of diversity of the small number of solutions in the pool while maintaining the quality of solutions \citep{marti2013multi,sorensen2006ma}. Then in the proposed metaheuristic, we try to manage the solutions in the pool by measuring its non-improving iterations. The detailed mechanism used in the proposed algorithm is shown in \Cref{algorithm:diversity-control}. The algorithm will re-generate solutions whenever $\vartheta$ exceeds the maximum limit of non-improving iterations, $\Theta_{\mathbb{E}}$.


            \begin{algorithm}
            	\footnotesize
            	\caption{Diversity control mechanism.}\label{algorithm:diversity-control}
            	\hspace*{\algorithmicindent} \textbf{input:} elite set $\mathbb{E}$, current best solution $S_{best}$, best solution before $S_0$\\
                \hspace*{\algorithmicindent} \qquad \quad improvement iteration counter $\vartheta$, CVRP instance $\mathbf{I}$
            	\begin{algorithmic}[1]
            		\Procedure{EliteSetManagement}{$\mathbb{E}$, $S_{best}$, $S_0$, $\vartheta$, $\mathbf{I}$}
                    \If{$S_{best} < S_0$}
                        \State $\vartheta \gets 0$
                        \State \Return $\mathbb{E}$
                    \Else
                        \State $\vartheta \gets \vartheta +1$
                        \If{$\vartheta > \Theta_{\mathbb{E}}$}
                            \State $\mathbb{E} \gets \varnothing$ \Comment{re-start elite set}
                            \State $\mathbb{E} \gets \Call{GeneratingInitialSolutions}{\mathbf{I}}$ \Comment{re-generate initial solutions}
                            \State $\vartheta \gets 0$
                            \State \Return $\mathbb{E}$
                        \EndIf
                    \EndIf
            		\EndProcedure
            	\end{algorithmic}
            \end{algorithm}

    \subsection{Neighborhood Improvement}
    \label{subsec:neighborhood-improvement}
        In the proposed algorithm, the neighborhood improvement processes are composed of two major steps: perturbation and local search improvement. The full mechanism of neighborhood search is shown in \Cref{algorithm:neighborhood-improvement}. The perturbation mechanism is done by destroying and reconstructing a set of tours in the solution. The local search improvement is implemented through various local search operators, which are categorized into two groups. Within each group, the local search operators are randomly ordered.
    
\begin{figure}[H]
    \begin{center}
        \includegraphics[width=0.92\textwidth]{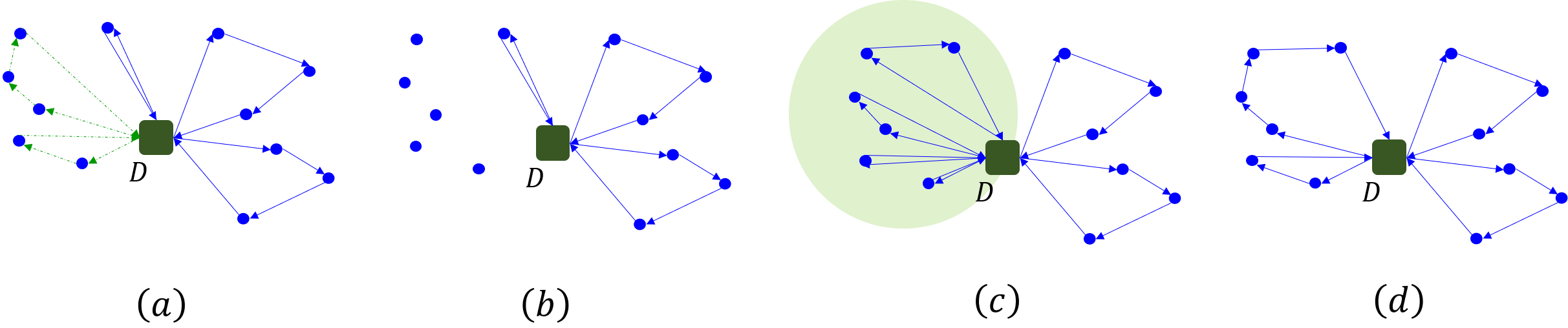}
        \caption{The pruned neighborhood improvement mechanism involves: (a) perturbing the solution by destroying a set of routes, (b) reinserting the destroyed customer nodes into existing or new routes, (c) applying local search focused on the recently moved customer nodes, and (d) generating a new solution.}
        \label{img:local-search-mechanims}
    \end{center}
\end{figure}
        \subsubsection{Local search improvement} 
        \label{subsubsec:local-search-improv}
            Local search performs well for improving solutions in metaheuristics \citep{arnold2019knowledge,Prodhon2016}, where iteratively refines a solution through small, local modifications called moves. In the proposed algorithm, we apply both intra and inter-route local search, where the intra improves nodes within a single route, while the inter operates across multiple routes. To enhance efficiency, local search is restricted to the perturbed regions, from operation line 5 in \Cref{algorithm:neighborhood-improvement}. This targeted search area (illustrated as the green region in \Cref{img:local-search-mechanims}(c)), has its maximum coverage defined in \Cref{table:paramater-setting}. The local searches used in the proposed algorithm are summarized below:
            
            \begin{itemize}
                \item \textbf{Relocate and Swap:} Relocate will insert a customer into a new position, while swap exchanges them. Both are applied within or between routes.
                \item \textbf{2-Opt and 2-Opt*:} 2-Opt removes two edges from a route and reconnects them to form a shorter path \citep{JUNGER1995225}. 2-Opt* extends this by exchanging sub-sequences between two routes.
                \item \textbf{CROSS-Exchange:} Removes four edges from two routes and reconnects them with four new edges \citep{taillard1997tabu}. 
                \item \textbf{Path-Move:} Moves a pair of consecutive customers within or between routes \citep{soto2017multiple}. 
                \item \textbf{Chain-Move:} Based on ejection chains \citep{glover1996ejection,rego2001node}, where begins with an infeasible relocation or path move, followed by a sequence of repairs until feasibility is restored or limits are reached.
            \end{itemize}
            
            \begin{algorithm}
            	\footnotesize
            	\caption{Neighborhood search.}\label{algorithm:neighborhood-improvement}
            	\hspace*{\algorithmicindent} \textbf{input:} elite set $\mathbb{E}$, current best solution $S_{best}$
            	\begin{algorithmic}[1]
            		\Procedure{NeighborhoodSearch}{$\mathbb{E},S_{best}$}
                    \For{$n \gets 1 \textbf{ to } \mathbb{E}_{min}$} 
                        \State $S \gets \arg \min_{s \in \mathbb{E}} \Call{Cost}{s}$ \Comment{best solution in current elite set}
                        \State $\mathbb{E} \gets \mathbb{E} \setminus {S}$
                        \State $S' \gets \Call{DestroyRepairTour}{S,S_{best}}$
                        \State $S'' \gets \Call{LocalSearchImprovement}{S',S_{best}}$
                        \State $\mathbb{E} \gets \Call{UpdateEliteSet}{S'',\mathbb{E}}$
                    \EndFor
                    \State \Return $\mathbb{E}$, $S_{best}$
            		\EndProcedure
            	\end{algorithmic}
            \end{algorithm}
            
    \subsection{Path Relinking} 
    \label{subsec:path-relininking}   
        In path relinking, the initial solution is improved toward the guiding solution \citep{ho2006path}, but in VRP, transforming multiple routes can be challenging. To address this, a solution representation as a Giant Tour (GT) combined with a split algorithm to convert it back to VRP was proposed \citep{prins2004simple}. In this research, we propose a hybrid of split and path relinking, transforming both initial and guiding solutions into GT, applying path relinking, then converting intermediate solutions with the split algorithm. This generates new solutions, improving quality and diversity in the pool $\mathbb{E}$. Full steps are shown in \Cref{algorithm:path-relinking}, starting with identifying restricted neighborhoods and conducting neighborhood search within $L_{pr}$. For further details regarding the proposed path relinking are described in \ref{sec:appendix-iteratively-neighborhood-search}.

        \begin{algorithm}
        	\footnotesize
        	\caption{Proposed hybrid split and path relinking.}\label{algorithm:path-relinking}
        	\hspace*{\algorithmicindent} \textbf{input:} elite set $\mathbb{E}$, current best solution $S_{best}$
        	\begin{algorithmic}[1]
        		\Procedure{Split-PathRelinking}{$\mathbb{E},S_{best}$}
        		\State $(S_i, S_g) \gets \Call{GetRandomParents}{\mathbb{E},S_{best}}$
        		\State $T_i \gets \Call{RandomConcatenation}{S_i}$\Comment{transform into giant tour}
                \State $T_g \gets \Call{RandomConcatenation}{S_g}$\Comment{transform into giant tour}
                \State $(\Delta_{pr}, L_{pr}) \gets \Call{GetRestrictedNeighborhood}{T_i, T_g}$
                \State $L_{pr}' \gets \Call{ReOrderingCustomerList}{L_{pr}}$
                \State $N_{pr} \gets \left( \nicefrac{\Delta_{pr}}{2} \right) \cdot \eta_{pr}$
                \State $(\mathbb{E},S_{best}) \gets \Call{EvaluateNeighborhood}{T_i,T_g,N_{pr},L_{pr}',\mathbb{E},S_{best}}$
                \State \Return $\mathbb{E}$, $S_{best}$
        		\EndProcedure
        	\end{algorithmic}
        \end{algorithm}
        
        \paragraph{\textbf{Truncated path relinking}} 
            \cite{Resendel2005} demonstrated that there tends to be a higher concentration of better solutions close to the initial solutions explored by path relinking. Additionally, we may reduce the computational time while still possible to obtain good solutions by adapting this mechanism. Performing the truncated path relinking mechanism when exploring the restricted neighborhood can be done by introducing $\eta_{pr}$ as the index defining the portion of the path to be explored,  where $0 < \eta_{pr} \leq 1$, which the best value is shown in \Cref{table:paramater-setting}. As we utilized $\eta_{pr}$,  instead of evaluating all $\Delta_{pr}$ restricted neighborhoods, we will use $N_{pr}$ (defined as $N_{pr}:=\left( \nicefrac{\Delta_{pr}}{2} \right) \cdot \eta_{pr}$) as the main loop for evaluating the restricted neighborhood.

\section{Hybridizing Metaheuristics with Feature-Based Guidance}
\label{sec:hybridizing-ml-mh}
    As hypothesized in \Cref{subsec:integrations-mhml}, we aim to leverage the most important features to derive rules for enhancing the performance of the baseline metaheuristic. Following the previously developed the metaheuristic baseline in \Cref{sec:development-metaheuristic}, this section presents how we construct a feature-based guidance that derived from our explainability learning model.
    
    \subsection{Guidance for Diversity Control} 
    \label{subsec:guidance-for-diversity-control}
        In \Cref{subsubsec:diversity-control}, we use the number of non-improving iterations, $\vartheta$, to determine whether the solution pool $\mathbb{E}$ should be regenerated. Meanwhile, as discussed in \Cref{subsec:xplainingp-learning-model}, solution quality can also be evaluated based on capacity utilization through features S19 and S20. Therefore, instead of relying solely on the non-improving solution $\varphi$, we also incorporate the capacity utilization value via the variable $\mathcal{C}$. This approach helps assess whether the pool $\mathbb{E}$ remains possible for further improvement through evaluating its capacity utilization. We define a composite metric $\mathcal{C}$ to evaluate pool quality. First, for each restart $t$, we compute:
        \begin{equation} \label{eq:c_1c_2}
        \hat{c_1}^t := \sum^{\mathbb{E}}_{\varepsilon=1} \text{S19}(s) \qquad \hat{c_2}^t := \sum^{\mathbb{E}}_{\varepsilon=1} \text{S20}(s)
        \end{equation}
        where $\mathbb{E} \in s_1,\dots,\varepsilon$, in which $\mathbb{E}_{min} \leq \varepsilon \leq \mathbb{E}_{max}$. Over $T$ restarts, we aggregate:
        \begin{equation} \label{eq:alpha_beta}
        \alpha := \sum^T_{t=0} \hat{c_1}^t \qquad \beta := \sum^T_{t=0} \hat{c_2}^t
        \end{equation}
        with $\alpha$ and $\beta$ representing cumulative S19 and S20, respectively. As shown in \Cref{img:beeswarm}, high $\alpha$ and low $\beta$ indicate better solutions. Finally, we compute the quality threshold $\mathcal{C}$ as:
        \begin{equation} \label{eq:guidance}
        \mathcal{C} := \lceil \mathbf{M} \cdot \mathbf{W} \rceil = \lceil \mathbf{M} \cdot \nicefrac{3}{2} \cdot \left(\alpha - \beta \right) \rceil
        \end{equation}
        where $\mathbf{M}$ is a constant value, defined in \Cref{table:paramater-setting}. It shows that $\hat{c_1}$ corresponds to feature S19, and $\hat{c_2}$ corresponds to feature S20, which operate on different scales. Although normalization could equalize their numerical ranges, our experiments showed that it diminished the discriminative power of the guidance mechanism by flattening their relative influence. For this reason, the current formulation relies on the raw feature values, which also results in a simpler calculation.
        \begin{algorithm}[htbp]
        	\footnotesize
        	\caption{Guided diversity control mechanism.}\label{algorithm:guided-diversity-control}
        	\hspace*{\algorithmicindent} \textbf{input:} elite set $\mathbb{E}$, current best solution $S_{best}$, best solution before $S_0$\\
                \hspace*{\algorithmicindent} \qquad \quad feature threshold $\mathbf{W}$, improvement iteration counter $\vartheta$, CVRP instance $\mathbf{I}$
        	\begin{algorithmic}[1]
        		\Procedure{GuidedEliteSetManagement}{$\mathbb{E}$, $S_{best}$, $S_0$, $\mathbf{W}$, $\vartheta$, $\mathbf{I}$}
                \If{$S_{best} < S_0$}
                    \State $\vartheta \gets 0$
                    \State \Return $\mathbb{E}$
                \Else
                    \State $\vartheta \gets \vartheta +1$
                    \If{$\vartheta > \mathcal{C}$}
                        \State $\mathbb{E} \gets \varnothing$ \Comment{re-start elite set}
                        \State $(\mathbb{E}, \alpha, \beta) \gets \Call{GeneratingInitialSolutions}{\mathbf{I}}$ \Comment{re-generate initial solutions}
                        \State $\mathbf{W} \gets \nicefrac{ \left( \mathbf{W} + \alpha + \beta \right) }{2}$
                        \State $\mathcal{C} \gets \lceil \mathbf{W} \cdot \mathbf{M} \rceil$\Comment{updating threshold}
                        \State $\vartheta \gets 0$
                        \State \Return $\mathbb{E}$
                    \EndIf
                \EndIf
        		\EndProcedure
        	\end{algorithmic}
        \end{algorithm}
        
    \subsection{Metaheuristic with Guidance for Diversity Control}
    \label{subsec:metaheuristic-with-guidance}
\begin{figure}[htbp]
    \begin{center}
        \includegraphics[width=0.88\textwidth]{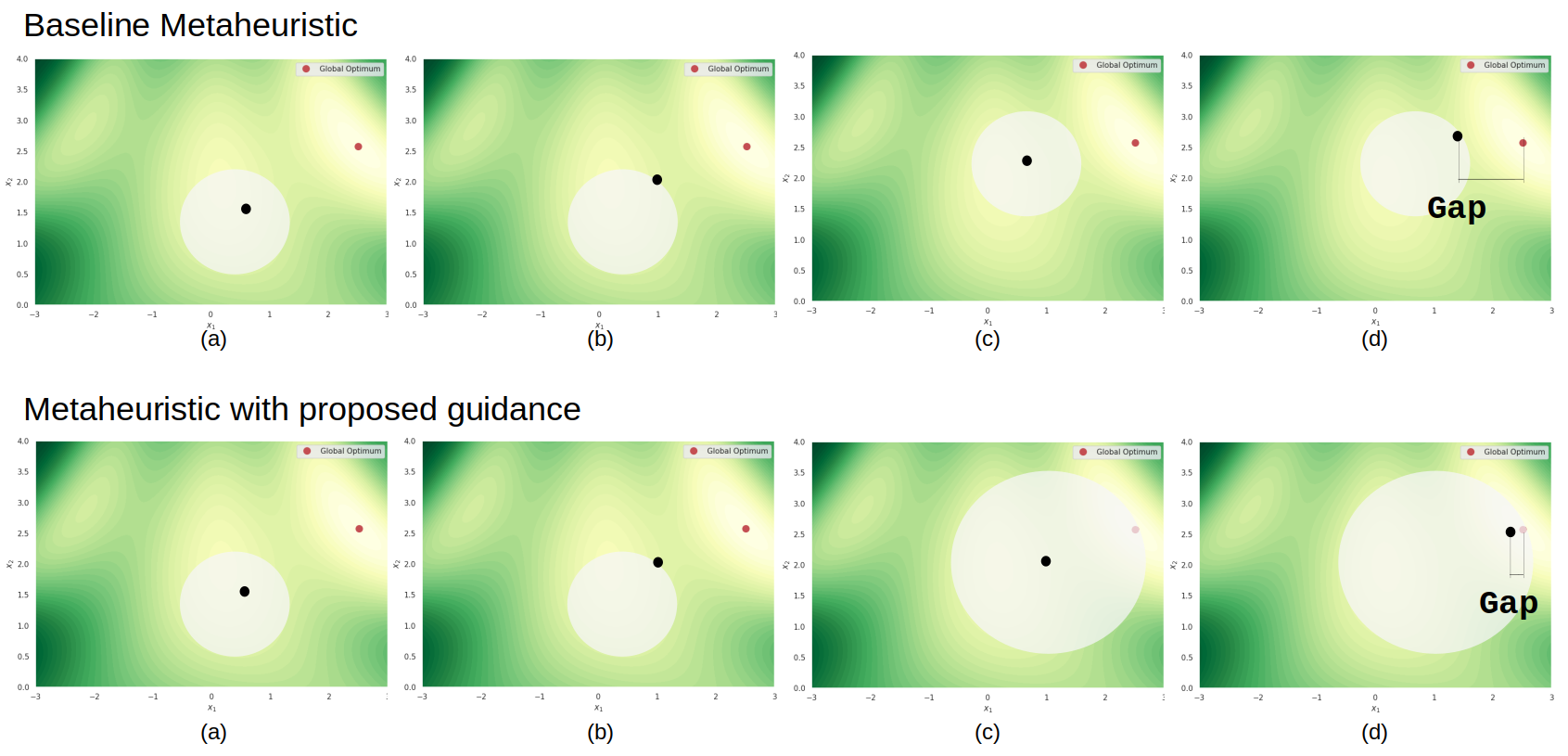}
        \caption{The illustration demonstrates the performance of the proposed guidance.}
        \label{img:illustration-guidance}
    \end{center}
\end{figure}

        As outlined in \Cref{eq:guidance}, the guidance mechanism $\mathbf{W}$ is implemented in \Cref{algorithm:guided-diversity-control}, and integrated into \Cref{algorithm:guided-ms-overview}. As shown in \Cref{img:illustration-guidance}, since the baseline uses a multi-start mechanism, it is typically invoked when the algorithm detects no further potential for improvement, \textit{i.e.}, reaching a local peak. As shown in (c) and (d) in \Cref{img:illustration-guidance}, the proposed guidance dynamically adjusts the search region (\textit{i.e.}, the timing for regenerating elite set $\mathbb{E}$), enabling the algorithm to better detect potential improvements beyond local valleys. In the next section, we conduct computational experiment to assess the performance of the guided version of the proposed algorithm \Cref{algorithm:guided-ms-overview} with the baseline \Cref{algorithm:ms-overview}. This evaluation aims to determine whether the guidance mechanism improves performance and supports our hypothesis in \Cref{subsec:integrations-mhml} regarding enhanced solution quality.

        \begin{algorithm}[htbp]
        	\footnotesize
        	\caption{Guided MS for solving CVRP.}\label{algorithm:guided-ms-overview}
        	\hspace*{\algorithmicindent} \textbf{input:} CVRP instance $\mathbf{I}$
        	\begin{algorithmic}[1]
        		\Procedure{Guided-MS-CVRP}{$\mathbf{I}$}
        		\State $(\mathbf{W},\mathbb{E}) \gets \Call{GeneratingInitialSolutions}{\mathbf{I}}$
                \State $S_{best} \gets \arg \min_{s \in \mathbb{E}} \Call{Cost}{s}$
                \State $\vartheta \gets 0$ \Comment{non-improving iteration counter}
                \Repeat
                    \State $(\mathbb{E},S_{best}) \gets \Call{NeighborhoodSearch}{\mathbb{E},S_{best}}$
                    \State $(\mathbb{E},S_{best}) \gets \Call{Split-PathRelinking}{\mathbb{E},S_{best}}$
                    \State $(\mathbf{W},\mathbb{E},S_{best}) \gets \Call{GuidedEliteSetManagement}{\mathbb{E},S_{best},S_0,\mathbf{W},\vartheta,\mathbf{I}}$
                \Until{$T_{max}$}
                \State \Return $S_{best}$
        		\EndProcedure
        	\end{algorithmic}
        \end{algorithm}
        
\section{Experiment and Analysis}
\label{sec:experiment-analysis}
    The algorithm was implemented in C++ and compiled using g++ 8.3.0. The experiment was performed on a 64-bit machine with dual Intel Xeon Gold 6130 CPUs and 192GB RAM, running Ubuntu 22.04.5 LTS. To account for randomness \citep{matsumoto1998mersenne}, for all experiments, each instance was run five times with distinct seeds, defined as the run counter minus one. Throughout the experimentation, we refer to the following:

    \begin{itemize}
        \item BKS: the total cost value of the best-known solutions. All the information related to the instances and best-known solutions are available at \url{https://galgos.inf.puc-rio.br/cvrplib/en/instances}.
        \item Gap: the relative difference between the obtained solution and the best-known solution, computed using \Cref{eq:gap-to-opt} by substituting the optimal value with the best-known solution.
    \end{itemize}

    The source code of the proposed algorithm can be downloaded from \url{https://github.com/bachtiarherdianto/MS-CVRP} alongside the instruction for replicating the experiment of the proposed feature-guided algorithms.

    \subsection{Parameter Tuning}
    \label{subsec:parmeter-tuning}
        The parameters used consist of parameters for pruning the Clarke and Wright when constructing a solution, parameters to control the size of the pool of elite set solution, and the search intensification. The details for these values are summarized in \Cref{table:paramater-setting}.

        \begin{table}[htbp]
            \caption{Parameters of the proposed algorithm.}
            \label{table:paramater-setting}
            \vspace*{0.1cm}
            \centering
            \scalebox{0.8}
            {
                \begin{tabular}{lcclrlr}
                \toprule
                \multicolumn{4}{l}{Parameter} & Value&&Described in\\
                \midrule
                $\mathcal{N}_{cw}$ &&& Size of saving table&100&&\Cref{subsec:generating-inisol}\\
                $\mathbb{E}_{max}$&&& Maximum size of pool elite solution&3&&\Cref{subsec:eliteset}\\
                $\mathbb{E}_{min}$ &&& Minimum size of pool elite solution &2&&\Cref{subsec:eliteset}\\
                $\eta_{pr}$ &&& Truncated index &0.4&&\Cref{subsec:path-relininking}\\
                $\mathbf{M}$ &&& Restart constant &4000&&\Cref{subsec:guidance-for-diversity-control}\\
                \bottomrule
                \end{tabular}
            }
        \end{table}

    \subsection{Computational Experiment with Baseline Algorithm} 
    \label{subsec:xml-result}
\begin{figure}[htbp]
    \begin{center}
        \includegraphics[width=0.4\textwidth]{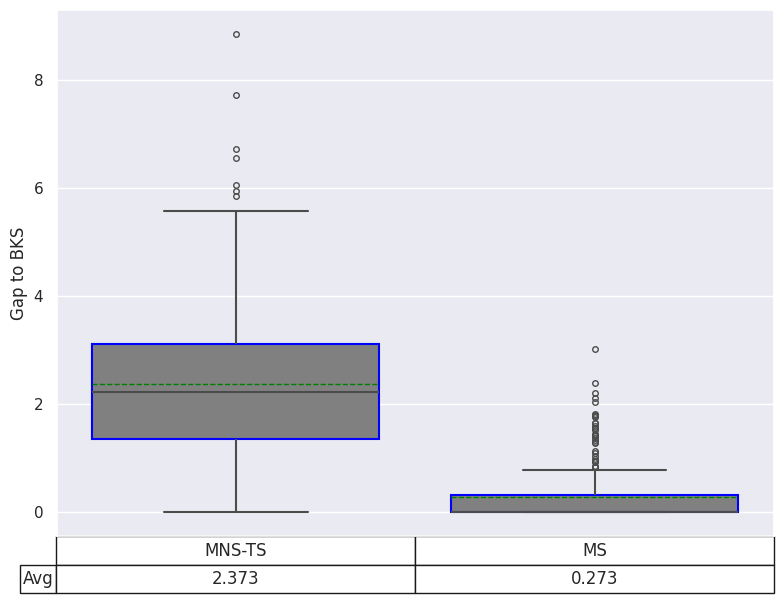}
        \caption{Boxplot comparing the baseline MNS-TS and the MS on subset $\mathbb{XML}$100 instances.}
        \label{img:boxplot-xml}
    \end{center}
\end{figure}

        To evaluate the effectiveness of the proposed guidance that applied in \Cref{algorithm:guided-ms-overview} compared with the baseline in \Cref{algorithm:ms-overview}, we conducted experiments on $250$ randomly selected $\mathbb{XML}$100 \citep{queiroga202110}. The MNS-TS algorithm, used to generate near-optimal solutions in \Cref{subsec:data-generation}, serves as the baseline benchmark. Each algorithm was given a $60$ second time limit per instance. The results are summarized in \Cref{img:boxplot-xml}, measuring the quality of the obtained solutions relative to the optimal values for each instance \citep{queiroga202110}. 
        
\begin{figure}[htbp]
    \begin{center}
        \includegraphics[width=0.4\textwidth]{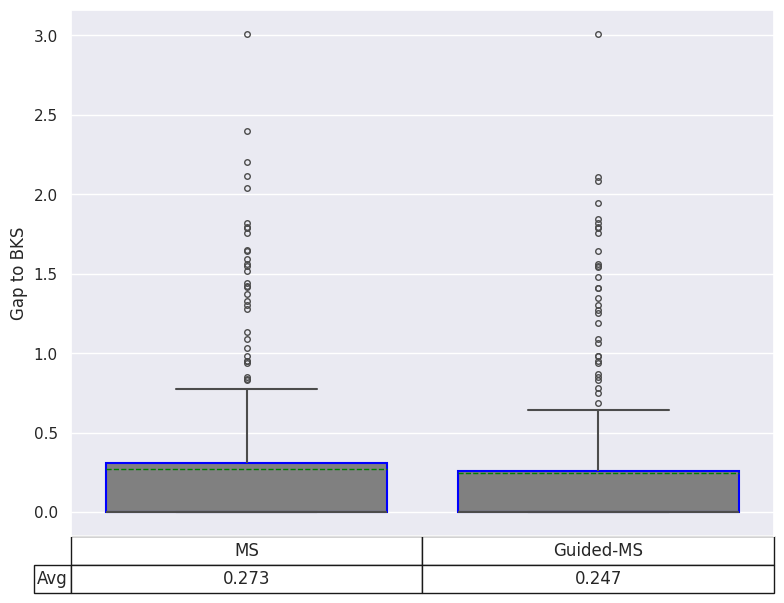}
        \caption{Boxplot the effect of the proposed guidance on the MS on subset $\mathbb{XML}$100 instances.}
        \label{img:boxplot-xml-ms}
    \end{center}
\end{figure}

        They show that all proposed algorithms outperform the baseline, with the guided version achieving the best performance. The comparison between MNS-TS and MS, is illustrated in \Cref{img:boxplot-xml} and MS and Guided-MS is illustrated in \Cref{img:boxplot-xml-ms}. Furthermore, a statistical tests are applied to validate the results shown in \Cref{img:boxplot-xml-ms}.
        
        \paragraph{\textbf{Statistical analysis of feature-based guidance}} 
        To assess the performance the proposed feature-based guidance on the baseline MS algorithm, we perform a statistical analysis. Hence, we performed the non-parametric one-tailed Wilcoxon signed-rank test \citep{JMLR:v7:demsar06a,arnold2021pils,accorsi2022guidelines,zarate2025machine}. The test hypotheses are defined in \Cref{hyp:a} and \Cref{hyp:b}.
        \setcounter{hyp}{-1}
        \begin{hyp} \label{hyp:a}
            \Call{AvgCost}{$\mathcal{S}'$} $\equiv$ \Call{AvgCost}{$\mathcal{S}$}
        \end{hyp}
        \begin{hyp} \label{hyp:b}
            \Call{AvgCost}{$\mathcal{S}'$} $<$ \Call{AvgCost}{$\mathcal{S}$}
        \end{hyp}

        In this research, we set $\alpha = 0.05$ \citep{arnold2021pils,zarate2025machine}. For \Cref{img:boxplot-xml-ms}, a one-tailed Wilcoxon signed-rank test yields $p$-value $= 0.00007$, indicating a statistically significant improvement. Thus, we reject \Cref{hyp:a} where the proposed guidance able to improve the baseline.

\begin{figure}[htbp]
    \begin{center}  
        \includegraphics[width=0.4\textwidth]{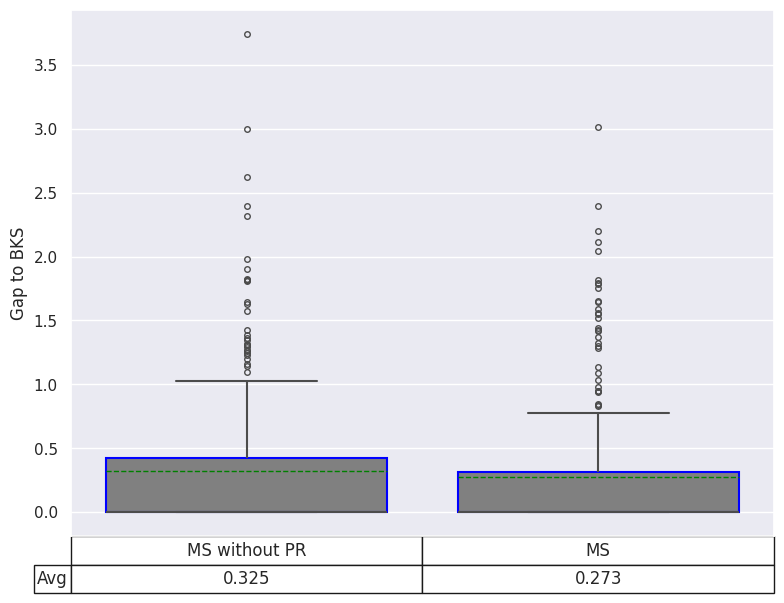} 
        \caption{Boxplot illustrating the contribution of the proposed path relinking mechanism.}
        \label{img:boxplot-pr-contribution}
    \end{center}
\end{figure}

        \paragraph{\textbf{Path relinking contributions}}
        We also measure the effectiveness of our proposed path relinking mechanism. \Cref{img:boxplot-pr-contribution} showing the effect of our proposed path relinking on the MS algorithm using $250$ randomly selected $\mathbb{XML}$100 instances. The comp
        We performed a one-tailed Wilcoxon signed-rank test ($\alpha = 0.05$) to assess the contribution of the proposed path relinking on the performance of the MS algorithm, as shown in \Cref{img:boxplot-pr-contribution}. \Cref{img:boxplot-pr-contribution}, confirming the significant contribution of the mechanism. For \Cref{img:boxplot-pr-contribution}, on $250$ $\mathbb{XML}$100 instances, the test rejects \Cref{hyp:a} for MS with proposed path relinking ($p$-value $= 0.00016$), indicating that the average results are statistically better.
        
\begin{table}[htbp]
    \begin{center}
        \vspace*{-0.2cm}
        \caption{Comparison of solution quality with $T_{max} = N \times \nicefrac{240}{100}$ seconds over 5 runs. The table presents computational results on the $\mathbb{X}$ instances \citep{UCHOA2017845}.}
        \vspace*{0.3cm}
        \setlength\tabcolsep{6pt}
        \label{table:result-stoa-summary}
        \scalebox{0.88}{
            \begin{tabular}{l r r r r r r}
                \toprule
                \textbf{Measurement} & \textbf{LKH-3} & \textbf{FILO} & \textbf{Hexaly} & \textbf{HGS} & \textcolor{blue}{\textbf{MS}} & \textcolor{blue}{\textbf{Guided-MS}} \\
                \midrule
                Average Gap & 1.029 & 0.368 & 1.731 & \textbf{0.120} & \textcolor{blue}{0.892} & \textcolor{blue}{0.8323} \\
                Median Gap & 0.898 & 0.351 & 1.120 & \textbf{0.059} & \textcolor{blue}{0.926} & \textcolor{blue}{0.8329} \\
                \bottomrule
            \end{tabular}
        }
    \end{center}
\end{table}

\begin{figure}[htbp] 
    \begin{center}
        \includegraphics[width=0.6\textwidth]{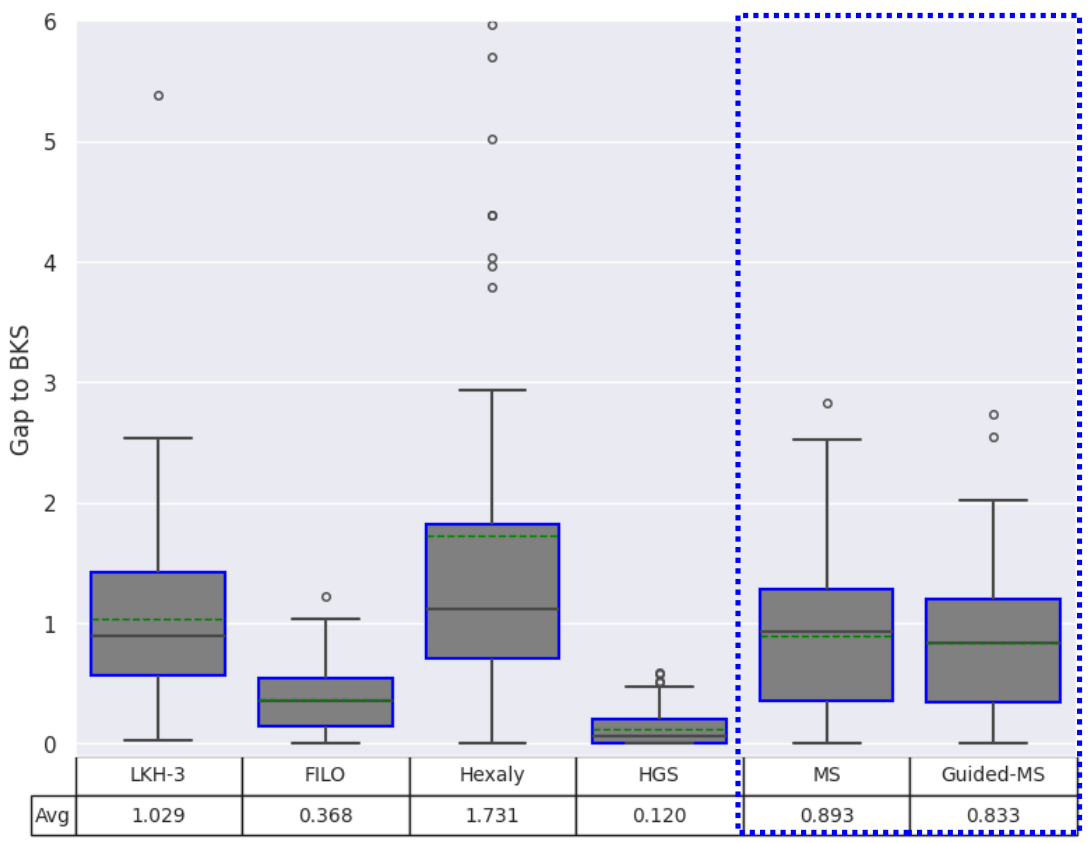}
        \caption{Summary the computational results on large instances.}
        \label{img:boxplot-stoa}
    \end{center}
\end{figure}

    \subsection{Testing the Proposed Guidance on Large-Scale Instances}
    \label{subsec:x-result}
\begin{figure}[htbp]
    \begin{center}
        \includegraphics[width=0.86\textwidth]{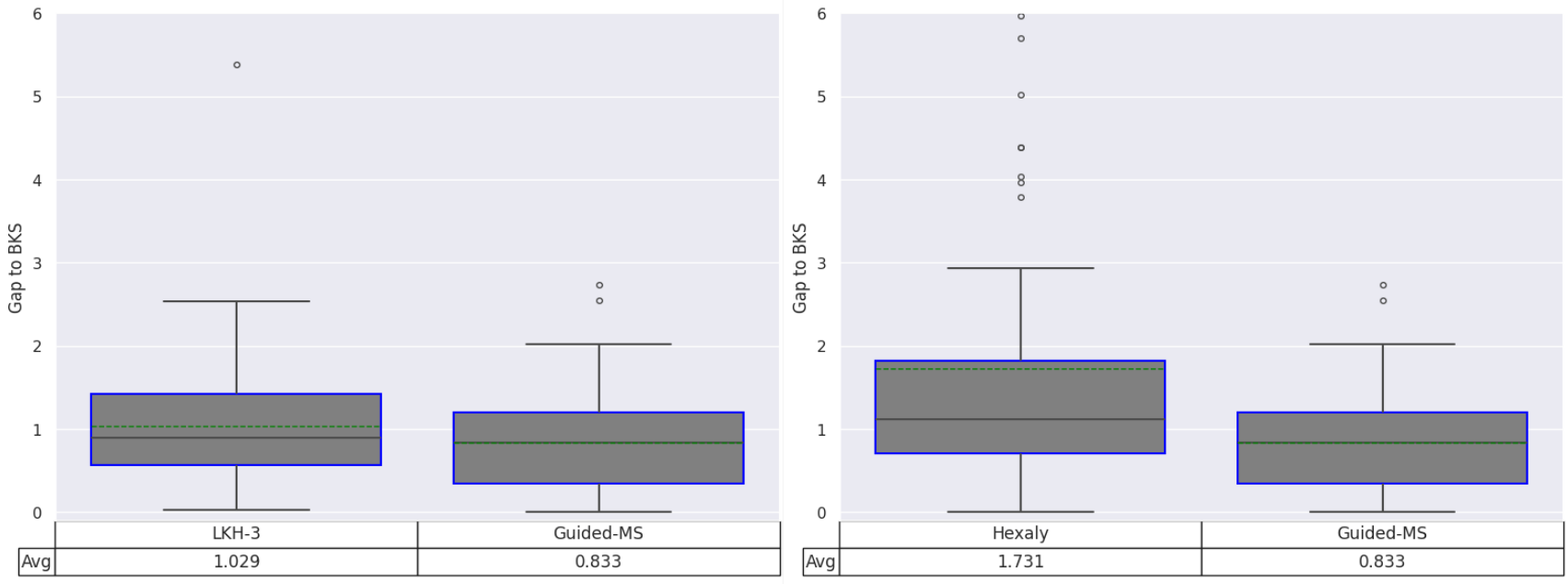}
        \caption{The comparison between Guided-MS, LKH-3, and Hexaly on the $\mathbb{X}$ instances.}
        \label{img:boxplot-x-paired}
    \end{center}
\end{figure}

        The results on the $\mathbb{X}$ instances \citep{UCHOA2017845} for the MS and Guided-MS algorithm are shown in \Cref{table:result-stoa-summary}. We compare its performance with other CVRP solvers: LKH-3\footnote{LKH-3.0.9 solver: \url{http://webhotel4.ruc.dk/~keld/research/LKH-3/}} \citep{helsgaun2017extension}, HGS\footnote{HGS solver: \url{https://github.com/vidalt/HGS-CVRP}} \citep{vidal2022hybrid}, and FILO\footnote{FILO solver: \url{https://github.com/acco93/filo}} \citep{accorsi2021fast}, and the commercial solver Hexaly\footnote{Hexaly solver 13.0: \url{https://www.hexaly.com}}. All algorithms were run under identical machine and stopping criteria ($T_{max} = N \times \nicefrac{240}{100}$ seconds). While the proposed metaheuristic does not outperform all benchmarks, it consistently outperforms LKH-3 and Hexaly. The comparison between Guided-MS, LKH-3, and Hexaly is also detailed in \Cref{img:boxplot-x-paired}. 

\begin{figure}[htbp]
    \begin{center}
        \includegraphics[width=0.46\textwidth]{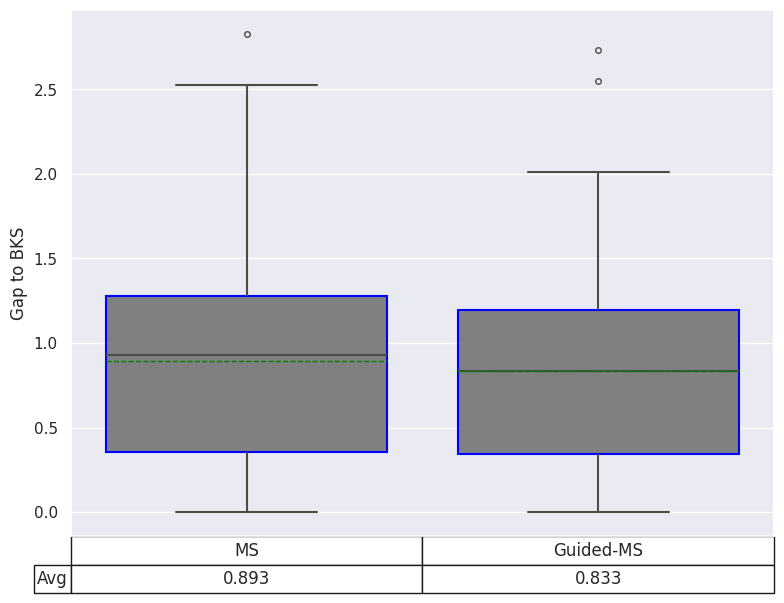}
        \caption{The effect of the proposed guidance on the $\mathbb{X}$ instances.}
        \label{img:boxplot-x-ms}
    \end{center}
\end{figure}

        To assess the scalability of the proposed feature-based guidance, \Cref{img:boxplot-x-ms} shows that Guided-MS outperforms the baseline MS algorithm. On the $\mathbb{X}$ instances, the test rejects \Cref{hyp:a} for Guided-MS ($p$-value $= 0.000027$), indicating that the proposed guidance provides a statistically significant improvement over the baseline. Furthermore, \Cref{img:boxplot-x-paired} illustrates the performance of Guided-MS compared to Hexaly (left) and LKH-3 (right). On the $\mathbb{X}$ instances, the test rejects \Cref{hyp:a} for Guided-MS, indicating statistically significant improvements over both Hexaly ($p$-value $= 1.07 \cdot 10^{-10}$) and LKH-3 ($p$-value $= 1.739 \cdot 10^{-5}$).

    \subsection{Generalization the Proposed Guidance}
    \label{subsec:generalization}
        As shown in \Cref{table:result-stoa-summary}, HGS performs very well compared to other state-of-the-art metaheuristic algorithms. Hence, to evaluate the generalization ability of the guidance, we integrated our proposed feature-based guidance into HGS \citep{vidal2022hybrid}, resulting in a variant referred to as HGS with guidance. This variant was tested on 30 sampled instances of the $\mathbb{X}$ instances \citep{UCHOA2017845}, using $T_{max} = N \times \nicefrac{240}{100}$ and 5 different random seeds.

        \paragraph{\textbf{Incorporating the guidance}}
        Unlike MS algorithm, which maintains diversity through a small elite set and multi-start mechanism (as described in \Cref{subsubsec:diversity-control}), HGS controls diversity through penalties and population balance between feasible and infeasible solutions \citep{vidal2022hybrid}. To incorporate the proposed guidance mechanism, we modified the restart mechanism of HGS, normally triggered after $\text{N}_{\text{IT}} = 20,000$ non-improving iterations. Following the principles in \Cref{subsec:integrations-mhml}, we dynamically adjust the restart threshold based on feature values from the current best solution $S_{best}$. Specifically, we redefine the maximum non improving iteration as:
        \begin{equation} \label{eq:guided-hgs}
            \text{N}_{\mathcal{C}} = \text{N}_{\text{IT}} \cdot \dfrac{\text{S19}(S_{best})}{\left( 1 - \text{S20}(S_{best}) \right)}
        \end{equation}
        Then, in the guided version, $\text{N}_{\mathcal{C}}$ will simply replace $\text{N}_{\text{IT}}$ to determine maximum number of not improving iteration when the algorithm solving an instance.

\begin{figure}[htbp]
    \begin{center}
        \includegraphics[width=0.46\textwidth]{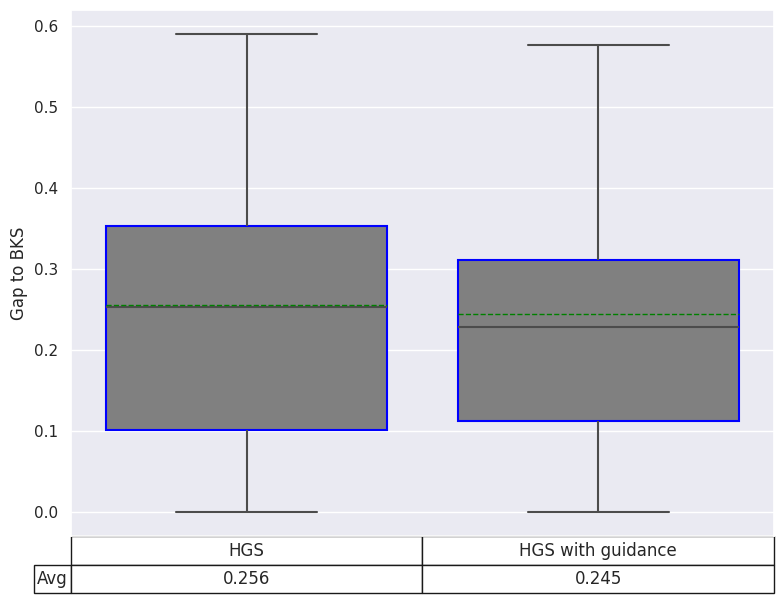}
        \caption{Summary of the computational results with baseline HGS.}
        \label{img:boxplot-generalization}
    \end{center}
\end{figure}

        \paragraph{\textbf{Experiment results}}
        As the baseline HGS already performs very good on small instances. Therefore, in this experiment, we tested the proposed guidance mechanism on $30$ $\mathbb{X}$ instances with size $n > 350$ to evaluate its effectiveness in enhancing performance on medium and large instances. As illustrated in \Cref{img:boxplot-generalization}, the proposed guidance successfully improves the performance of the baseline HGS. On $\mathbb{X}$ instances, the test rejects \Cref{hyp:a} for HGS with guidance ($p$-value $= 0.014$), indicating that the proposed guidance is statistically improving the baseline HGS algorithm.

    \subsection{Result on Very Large-Scale Instances}
    \label{subsec:result-very-large}
\begin{figure}[htbp]
    \begin{center}
        \includegraphics[width=1\textwidth]{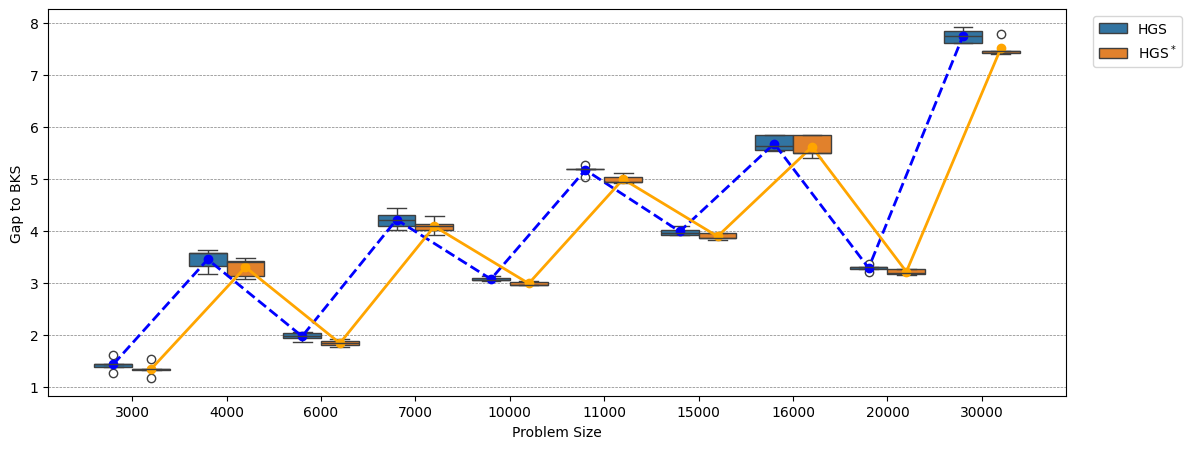}
        \caption{Effect of proposed guidance on HGS in solving $\mathbb{B}$ instances.}
        \label{img:hgs-line-b}
    \end{center}
\end{figure}

        To further assess the scalability of the proposed feature-based guidance, we conduct experiments using a set of very large-scale instances from the $\mathbb{B}$ set \citep{ARNOLD201932}. The summary of the experimental results using a set of very large-scale instances from the $\mathbb{B}$ set are shown in \Cref{img:hgs-line-b}. 

\begin{figure}[htbp]
    \begin{center}
        \includegraphics[width=0.66\textwidth]{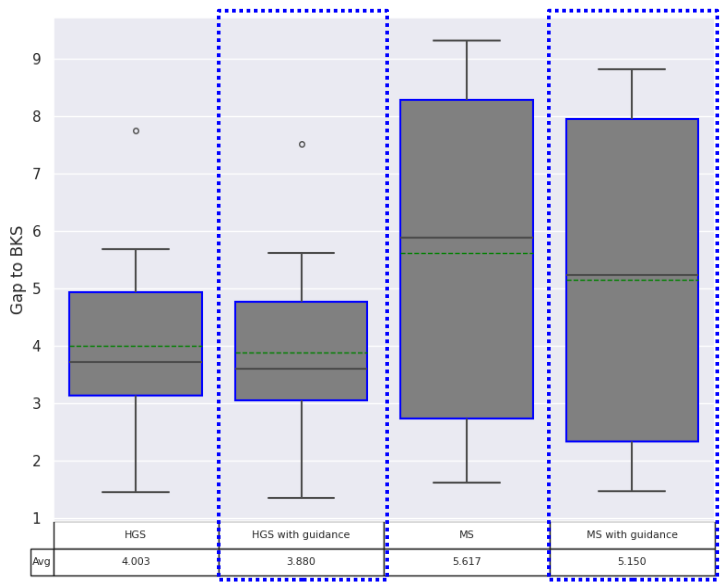}
        \caption{Effect of the proposed guidance on very large instances.}
        \label{img:boxplot-b} 
    \end{center}
\end{figure}

        As illustrated in \Cref{img:boxplot-b}, the proposed guidance can outperform both baseline algorithms. Furthermore, on 10 instances from the $\mathbb{B}$ set, the test rejects \Cref{hyp:a} for Guided-MS ($p$-value $= 0.001$) and for HGS with guidance ($p$-value $= 0.001$), suggesting that the proposed guidance statistically improves the baseline algorithms when solving $\mathbb{B}$ instances.

\section{Conclusion}
\label{sec:conclusion}
    In this paper, we propose a feature-based guidance mechanism to enhance the performance of metaheuristic algorithms for solving the CVRP. Using $10,000$ instances \citep{queiroga202110}, we generate a dataset comprising both optimal and near-optimal solutions, where optimal solutions are taken from the dataset and near-optimal ones are obtained using the MNS-TS algorithm \citep{soto2017multiple,lucas2020reducing}. A set of instance and solution features is used to train classification models, and SHAP analysis is leveraged to interpret and evaluate feature behavior. The results from this analysis form the basis for formulating the guidance mechanism.

    To initially implement the proposed guidance, we design a new metaheuristic that integrates our proposed novel path-relinking mechanism. This new mechanism is statistically proven to improve the performance of the algorithm. Furthermore, experimental results demonstrate that the proposed feature-based guidance capable to enhances the performance of the baseline metaheuristic. To evaluate the generalization capability and scalability of the proposed guidance, we also conducted a series of experiments and observed statistically significant improvements across all settings.
    
\newpage
\appendix

\section{Appendix}

\subsection{Feature of VRP}
\label{sec:appendix-feature-vrp}
    \paragraph{\textbf{Mathematical notations for features of VRP}} 
        In the context of VRP features, let $R$ denote the set of all routes in a solution. Each route $r_k = \{c_1, c_2, ..., c_k\} \in R$ is a sequence of customer nodes. The neighborhood rank between nodes $c_i$ and $c_j$ is $\mathcal{R}_{ij}$, and the average neighborhood rank for route $k$ is defined as $\mathcal{R}_k = \nicefrac{ \left( \sum_{i,j \in k} \mathcal{R}_{ij} \right) }{k}$. Each route $k$ consists of edges $(D,c_1)$, $(c_1,c_2)$, $\dots$, $(c_k,D) \in E$, where $D$ is the depot. The Euclidean distance between nodes $c_j$ and $c_k$ is $d(c_j,c_k)$, and $x(c_k), y(c_k)$ denote the coordinates of node $c_k$. The demand at node $c_j$ is $q(c_j)$. The coordinates of the center of gravity $G_k$ of route $k$ are computed as $x(G_k)= \nicefrac{ \left( \sum x(c_k)+x(D) \right)}{k+1}$ and $y(G_k)= \nicefrac{ \left( \sum y(c_k)+y(D) \right)}{k+1}$ \citep{arnold2019makes}. 
        
        Let $L_{G_k}$ be the line from $D$ to $G_k$, and $d(L_{G_k}, c_i)$ the signed distance of customer $c_i$ to this line (positive if on the right, negative otherwise). The angular difference between nodes $c_j$ and $c_k$ with respect to $D$ is $\text{rad}(c_j, c_k)$. The number of route intersections between $r_1$ and $r_2$ is $I(r_1, r_2)$. 

    \paragraph{\textbf{Instance features}}
        Here are the specific details about features that depend on the respective instance. These features are drawn from previous research \citep{arnold2019makes, lucas2020reducing}.
        
    \begin{justify}
        \textbf{I01: } Number of customers
        \begin{equation} \label{eq:i01}
            \text{I01}(S) := N = V - 1
        \end{equation}
        \textbf{I02: } Number of vehicles
        \begin{equation} \label{eq:i02}
            \text{I02}(S) := R
        \end{equation}
        \textbf{I03: } Degree of capacity utilization
        \begin{equation} \label{eq:i03}
            \text{I03}(S) := \dfrac{\sum_{j \in N} q(c_j)}{Q \cdot R}
        \end{equation}
        \textbf{I04: } Average distance between each pair of customers
        \begin{equation} \label{eq:i04}
            \text{I04}(S) := \dfrac{\sum_{i,j \in N \setminus i=j}d(c_i,c_j)}{N}
        \end{equation}
        \textbf{I05: } Standard deviation of the pairwise distance between customers 
        \begin{equation} \label{eq:i05}
            \text{I05}(S) := \sqrt{\dfrac{\sum_{i,j \in N \setminus i=j}(d(c_i,c_j)-\text{I04})^2}{N}}
        \end{equation}
        \textbf{I06: } Average distance from customers to the depot
        \begin{equation} \label{eq:i06}
            \text{I06}(S) := \dfrac{\sum_{i \in N}d(c_i,D)}{N}
        \end{equation}
        \textbf{I07: } Standard deviation of the distance from customers to the depot
        \begin{equation} \label{eq:i07}
            \text{I07}(S) := \sqrt{\dfrac{\sum_{i \in N}(d(c_i,D)-\text{I06})^2}{N}}
        \end{equation}
        \textbf{I08: } Average radians of customers towards the depot
        \begin{equation} \label{eq:i08}
            \text{I08}(S) := \dfrac{\sum_{i \in N}\text{rad}(c_i,D)}{N}
        \end{equation}
        \textbf{I09: } Standard deviation of the radians of customers towards the depot
        \begin{equation} \label{eq:i09}
            \text{I09}(S) := \sqrt{\dfrac{\sum_{i \in N}(\text{rad}(c_i,D)-\text{I08})^2}{N}}
        \end{equation}
    \end{justify}

\begin{figure}[htbp]
    \begin{center}
        \includegraphics[width = 0.42 \textwidth]{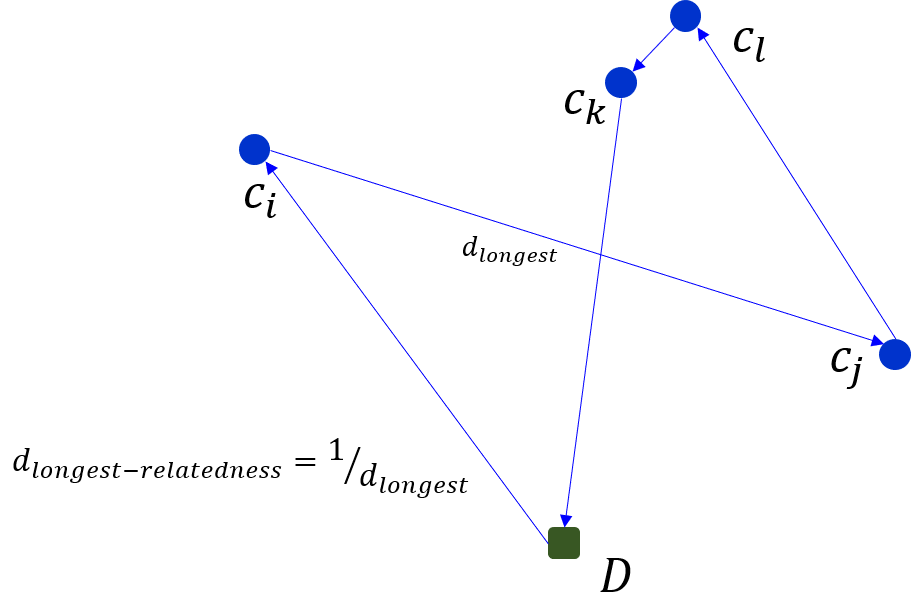}
        \caption{Illustration of longest-relatedness distance in a route of a solution.}
        \label{img:longest-relatedness}
    \end{center}
\end{figure}

    \paragraph{\textbf{Solution features}}
        These features, mostly adapted from previous studies \citep{arnold2019makes, lucas2020reducing}, capture structural properties of solutions. We also propose additional features, including those related to capacity utilization and longest-route metrics (illustrated in \Cref{img:longest-relatedness}), resulting in 22 solution features in total.
    
    \begin{justify}
    
        \textbf{S01: } Average width of each route
        \begin{equation} \label{eq:s01}
            \text{S01}(S) := \dfrac{1}{R} \cdot \sum_{k \in R} \left( \max_{j \in k} d(L_{G_k}, c_j) - \min_{j \in k} d(L_{G_k}, c_j) \right)
        \end{equation}
        \textbf{S02: } Standard deviation of the width of each route
        \begin{equation} \label{eq:s02}
            \text{S02}(S) := \sqrt{\dfrac{\sum_{k \in R} ((\max_{j \in k} d(L_{G_k}, c_j) - \min_{j \in k} d(L_{G_k}, c_j)) - \text{S01})^2}{R}}
        \end{equation}
        \textbf{S03: } Average span of each route
        \begin{equation} \label{eq:s03}
            \text{S03}(S) := \dfrac{1}{R} \cdot \sum_{k \in R} \max_{i,j \in k} \text{rad}(c_i, c_j)
        \end{equation}
        \textbf{S04: } Standard deviation of the span of each route
        \begin{equation} \label{eq:s04}
            \text{S04}(S) := \sqrt{\dfrac{\sum_{k \in R} (\max_{i,j \in k} \text{rad}(c_i, c_j) - \text{S03})^2}{R}}
        \end{equation}
        \textbf{S05: } Average depth of each route
        \begin{equation} \label{eq:s05}
            \text{S05}(S) := \dfrac{1}{R} \cdot \sum_{k \in R} \max_{j \in k} d(c_j, D)
        \end{equation}
        \textbf{S06: } Standard deviation of the depth of each route
        \begin{equation} \label{eq:s06}
            \text{S06}(S) := \sqrt{\dfrac{\sum_{k \in R} (\max_{j \in j} d(c_j, D) - \text{S05})^2}{R}}
        \end{equation}
        \textbf{S07: } Average of the distance of the first and last edge of each route divided by the total length of the route
        \begin{equation} \label{eq:s07}
            \text{S07}(S) := \dfrac{1}{2 \cdot R} \cdot \sum_{k \in R} \dfrac{d(D,c_1)+d(c_k,D)}{\sum_{i,j \in k} d(c_i, c_j)}
        \end{equation}
        \textbf{S08: } Mean length of the longest edge of each route
        \begin{equation} \label{eq:s08}
            \text{S08}(S) := \dfrac{1}{R} \cdot \sum_{k \in R} \max_{i,j \in k} d(c_i,c_j)
        \end{equation}
        \textbf{S09: } Length of the longest edge of all each route, divided by the average of the length of each route
        \begin{equation} \label{eq:s09}
            \text{S09}(S) := \dfrac{\max_{k \in R} \max_{i,j \in k} d(c_j,c_k)}{\nicefrac{ \left( \sum_{k \in R}  \sum_{i,j \in k }d(c_i,c_j) \right) }{R}}
        \end{equation}
        \textbf{S10: } Length of the longest interior edge of each route divided by mean of the length of each route
        \begin{equation} \label{eq:s10}
            \text{S10}(S) := \dfrac{\max_{k \in R} \max_{i,j \in k-1} d(c_{i+1},c_k)}{\sum_{k \in R} \sum_{i,j \in k}d(c_i,c_j)/{R}}
        \end{equation}
        \textbf{S11: } Mean length of the first and last edges of each route
        \begin{equation} \label{eq:s11}
            \text{S11}(S) := \dfrac{1}{2 \cdot R} \cdot \sum_{k \in R} (d(D,c_1)+d(c_k,D))
        \end{equation}
        \textbf{S12: } Average of demand of the first and last customer of each route
        \begin{equation} \label{eq:s12}
            \text{S12}(S) := \dfrac{1}{2 \cdot R} \cdot \sum_{k \in R} (q(c_1)+q(c_k))
        \end{equation}
        \textbf{S13: } Average of demand of the farthest customer
        \begin{equation} \label{eq:s13}
            \text{S13}(S) := \dfrac{1}{R} \cdot \sum_{k \in R} \max_{j \in k} q(c_j)
        \end{equation}
        \textbf{S14: } Standard deviation of the demand of the farthest customer
        \begin{equation} \label{eq:s14}
            \text{S14}(S) := \sqrt{\dfrac{\sum_{k \in R} \max_{j \in k} q(c_j) - \text{S13})^2}{R}}
        \end{equation}
        \textbf{S15: } Standard deviation of the length of each route
        \begin{equation} \label{eq:s15}
            \text{S15}(S) := \sqrt{\dfrac{\sum_{k \in R} (\sum_{i,j \in k}d(c_i,c_j) - (\sum_{k \in R} \sum_{i,j \in k} \nicefrac{d(c_i,c_j)}{R}))^2}{R}}
        \end{equation}
        \textbf{S16: } Mean distance between each route from their centre of gravity
        \begin{equation} \label{eq:s16}
            \text{S16}(S) := \dfrac{1}{R \cdot (R-1)} \cdot \sum_{r_1 \in R} \sum_{r_2 \in R \setminus r_1} d(G_{r_1}, G_{r_2})
        \end{equation}
        \textbf{S17: } Standard deviation of the number of customers of each route
        \begin{equation} \label{eq:s17}
            \text{S17}(S) := \sqrt{\dfrac{\sum_{k \in R} (k - \dfrac{V}{R})^2}{R}}
        \end{equation}
        \textbf{S18: } Average of the degree of the neighborhood for every route
        \begin{equation} \label{eq:s18}
            \text{S18}(S) := \text{S18}=\dfrac{1}{R} \cdot \sum_{k \in R} \mathcal{R}_k
        \end{equation}
        \textbf{S19: } The average of the capacity utilization for every route
        \begin{equation} \label{eq:s19}
            \text{S19}(S) := \dfrac{1}{R} \cdot \sum_{k \in R} \sum_{j \in k} q(c_j)/Q
        \end{equation}
        \textbf{S20: } Standard deviation of the capacity utilization for every route
        \begin{equation} \label{eq:s20}
            \text{S20}(S) := \sqrt{\dfrac{ \sum_{k \in R} \left( \sum_{j \in k} q(c_j)/Q - \left( \sum_{k \in R} \sum_{j \in k} q(c_j)/Q \right) \right)^2}{R}}
        \end{equation}
        \textbf{S21: } Average length of the longest distance-relatedness
        \begin{equation} \label{eq:s21}
            \text{S21}(S) := \dfrac{1}{R} \cdot \sum_{k \in R} \dfrac{1}{\max_{i,j \in k} d(c_i,c_j)}
        \end{equation}
        \textbf{S22: } Standard deviation of the length of the longest distance-relatedness
        \begin{equation} \label{eq:s22}
            \text{S22}(S) := \sqrt{ \dfrac{ \sum_{k \in R} \left( \nicefrac{1}{\max_{i,j \in k} d(c_i,c_j)} - \left( \sum_{k \in R} \nicefrac{\nicefrac{1}{\max_{i,j \in k} d(c_i,c_j)}}{R} \right) \right)^2 }{R}}
        \end{equation}
        
    \end{justify}

\newpage

\subsection{Feature Engineering} 
\label{sec:feature-engineering}
    Feature engineering, or feature discovery, involves extracting meaningful attributes from raw data to support model training \citep{hastie2009elements}. As described in \Cref{subsec:data-generation}, our dataset consists of $20,000$ data points that equally split between optimal and near-optimal solutions, generated by solving all 10,000 $\mathbb{XML}$100 instances \citep{queiroga202110}. These instances share a common size ($100$ customer nodes) but vary in depot position, customer distribution, demand patterns, and route characteristics, grouped into $378$ categories.
    
\begin{figure}[H]
    \begin{center}
        \includegraphics[width = 0.68\textwidth]{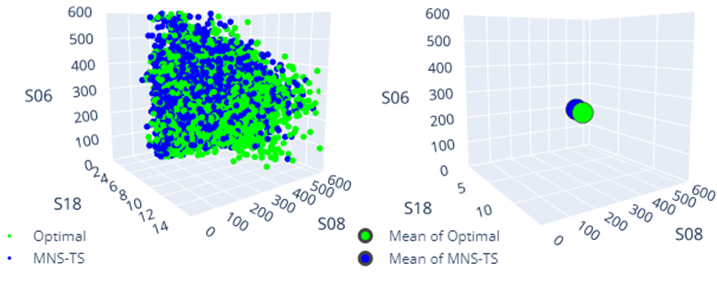}
        \caption{Illustration in the simplified 3D features space.}
        \label{img:data}
    \end{center}
\end{figure}
    
\begin{figure}[H]
    \begin{center}
        \includegraphics[width = 0.6\textwidth]{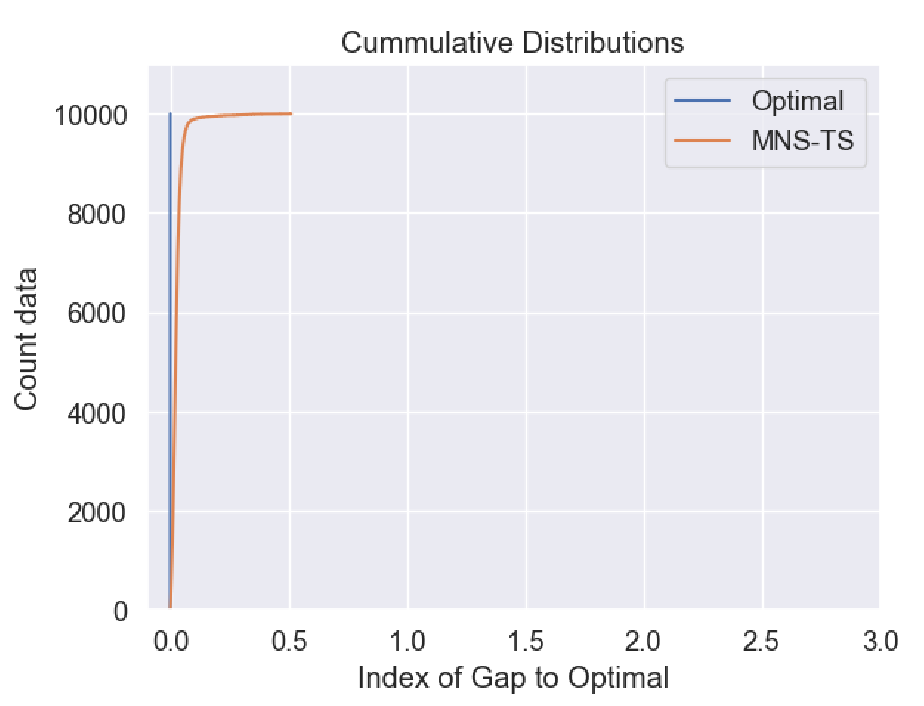}
        \caption{Cumulative distribution of optimal and near-optimal dataset.}
        \label{img:cumulative}
    \end{center}
\end{figure}

\begin{figure}[H] 
    \begin{center}
        \includegraphics[width = 0.98\textwidth]{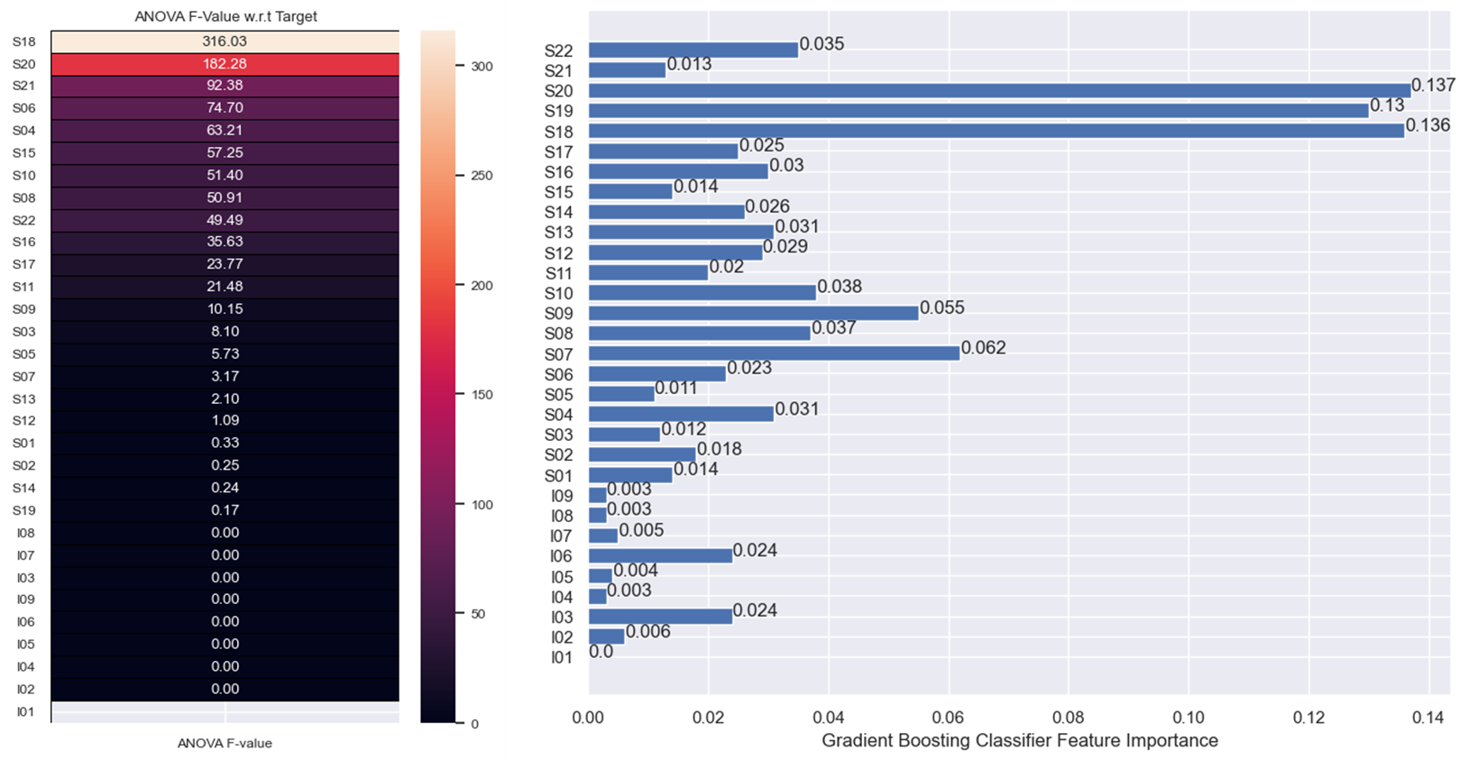}
        \caption{The order of ANOVA F-value (left) and impurity feature importance (right).}
        \label{img:feature}
    \end{center}
\end{figure}

    Features used for classification are grouped into instance and solution features, detailed in \ref{sec:appendix-feature-vrp}. As shown in \Cref{table:f1score}, the model accuracy remains modest due to the close proximity between optimal and near-optimal solutions. This is illustrated in the scatter plot of features S06, S08, and S18 (\Cref{img:data}) and further supported by the narrow gap representing the low diversity in solution quality, as shown in \Cref{img:cumulative}. Thus, $F_1$-score value in \Cref{table:f1score} show Gradient Boosting achieved the highest value, where direct feature importance from this model, based on impurity reduction (\Cref{img:feature}, right). It can used directly to identifies key features but ignores interactions among all features. 

\begin{figure}[H]
    \begin{center}
        \includegraphics[width = 0.88\textwidth]{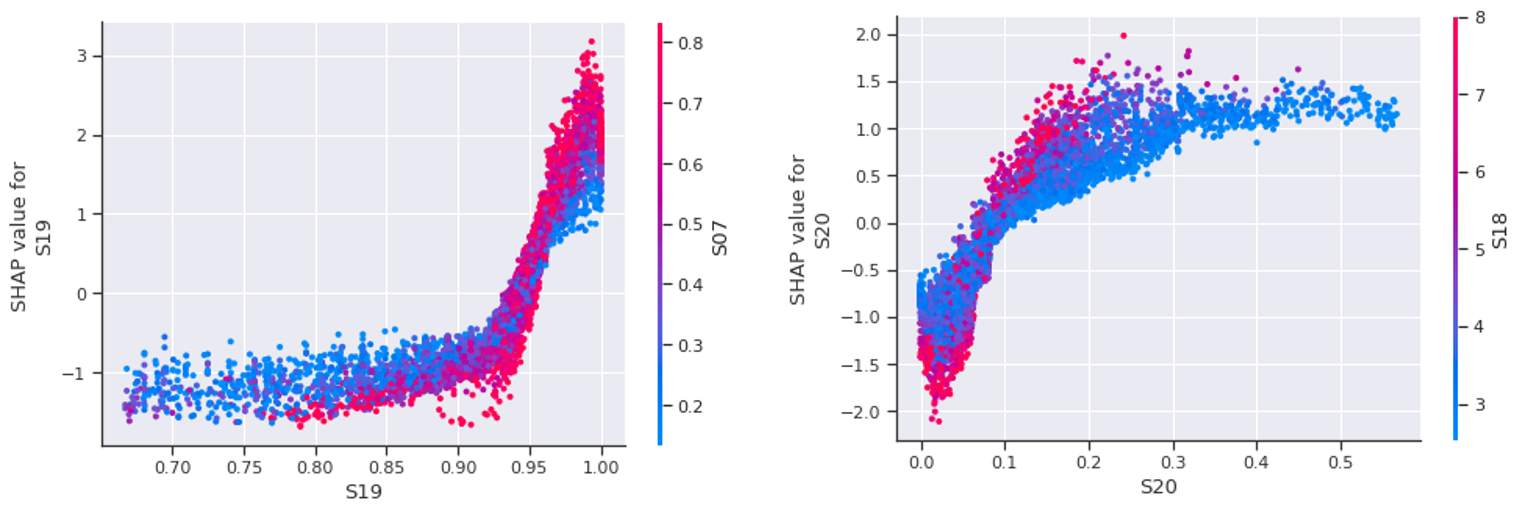}
        \caption{The dependency plot between S07 with S19 (left), and S20 with S18 (right).}
        \label{img:dependency}
    \end{center}
\end{figure}

    Meanwhile, by quickly analyzing how these features correlated with the quality of solutions using ANOVA F-values (\Cref{img:feature}, left), S18 and S20 appears statistically significant, but still this method does not account for feature interactions. Thus, as it does not capture the full range of relationships between features and the target variable, we still cannot conclude that S18 is the most important feature of the target variable. In contrast, the SHAP values presented in \Cref{img:beeswarm} provide a more comprehensive assessment by capturing both the individual contributions of each feature and their interaction. These interactions are visualized in the dependency plots in \Cref{img:dependency}. Although features such as S07 and S18 show secondary effects, their influence is conveyed through S19 and S20 rather than acting independently. This indicates that the contribution of additional features is largely captured by the interaction patterns already represented by the selected pair. As a result, adding more features is unlikely to yield meaningful benefit while introducing unnecessary complexity and computational cost.

\newpage

\subsection{Concatenation Method}
\label{sec:appendix-concat}
\begin{figure}[H] 
    \begin{center}
        \includegraphics[width=0.45\textwidth]{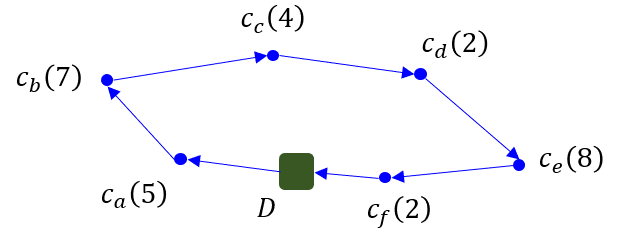} 
        \caption{Example of giant tour of VRP.}
        \label{img:giant-tour}
    \end{center}
\end{figure}

    \cite{prins2004simple} shows a new idea approach for solving the CVRP, called \textit{"route-first cluster-second"} paradigm. The approach is started by forming a giant tour (refer to \Cref{img:giant-tour}).
    
    The giant tour $T$ of solution $S$ is formed by using a simple randomized concatenation. The process starts by identifying the head customer of every route $r \in R$. Thus, continued by randomized concatenate the route.
    
    \begin{algorithm}
    	\footnotesize
    	\caption{Randomized concatenation mechanism.}\label{algorithm:concat}
    	\hspace*{\algorithmicindent} \textbf{input:} solution $S$
    	\begin{algorithmic}[1]
    		\Procedure{Concatenation}{$S$}
            \State $\mathbb{L}_{head} \gets \varnothing$ \Comment{list of first customer}
            \For{$r \in R$}
                \State $\mathbb{L}_{head}[r] \gets \Call{GetFirstCustomerOfRoute}{r}$
            \EndFor
            \State $\mathbb{L}_{head} \gets \Call{RandomReOrdering}{\mathbb{L}_{head}}$
            \State $T \gets \varnothing$ \Comment{initialize a giant tour}
            \For{$c \in \mathbb{L}_{head}$}
                \State $r \gets \Call{GetCustomerListFrom}{c}$ \Comment{get the route}
                \State $T \gets T \cup r$ \Comment{append route in to the giant tour}
            \EndFor
            \State \Return $T$
    		\EndProcedure
    	\end{algorithmic}
    \end{algorithm}

    In process, several intermediate giant tour solutions $T_{eval}$ are generated (see \ref{sec:appendix-iteratively-neighborhood-search}). If an intermediate solution is sufficient (see \Cref{algorithm:iteratively-neighborhood-eval}), it is transformed into a VRP solution via split algorithm \citep{prins2004simple}.

\newpage

\subsection{Neighborhood Search for Path Relinking} 
\label{sec:appendix-iteratively-neighborhood-search}
    To generate intermediate solutions to explore improvements, we need to explore the solution space between the initial and guiding solutions. To explore this solution space, we perform a neighborhood search between the initial solution and the guiding solution, as shown in \Cref{algorithm:path-relinking}, line 7. This neighborhood search process involves systematically examining neighboring solutions by making small modifications to the current solutions. An overview of the neighborhood search in the proposed path relinking is shown in \Cref{algorithm:iteratively-neighborhood-eval}.
    
    \begin{algorithm}
    	\footnotesize
    	\caption{Iteratively neighborhood evaluation processes.}\label{algorithm:iteratively-neighborhood-eval}
    	\hspace*{\algorithmicindent} \textbf{input:} initial solution $T_i$, guiding solution $T_g$, number of loop $N_{pr}$\\
        \hspace*{\algorithmicindent} \qquad \quad list of restricted neighborhood $L{pr}$, elite set $\mathbb{E}$, current best solution $S_{best}$
    	\begin{algorithmic}[1]
    		\Procedure{EvaluateNeighborhood}{$T_i$, $T_g$, $N_{pr}$, $L_{pr}$, $\mathbb{E}$, $S_{best}$}
            \State $T \gets T_i$
            \State $L_{tabu} \gets \varnothing$ \Comment{initialize tabu list}
            \For{$move \gets 1 \textbf{ to } N_{pr}$} 
                \State $(c_{swap},p_i,p_j) \gets \Call{GetPositionSwap}{T,T_g,L_{pr},L_{tabu}}$
                \State $T_{eval} \gets \Call{Swap}{T,p_i,p_j}$\Comment{intermediate solution}
                \If{$\Call{Cost}{T_{eval}} < \Call{Cost}{T}$}
                    \State $S_{eval} \gets \Call{Split}{T_{eval}}$\Comment{transform into VRP solution}
                    \State $\mathbb{E} \gets \Call{UpdateEliteSet}{S_{eval},\mathbb{E}}$
                    \If{$\Call{Cost}{S_{eval}} < \Call{Cost}{S_{best}}$}
                        \State $S_{best} \gets S_{eval}$
                    \EndIf
                \EndIf
                \State $L_{tabu} \gets L_{tabu} \cup c_{swap}$ \Comment{update tabu list}
            \EndFor
            \State \Return $\mathbb{E}$, $S_{best}$
    		\EndProcedure
    	\end{algorithmic}
    \end{algorithm}

    \begin{figure}[H]
    	\begin{center}
    		\includegraphics[width = 0.8\textwidth]{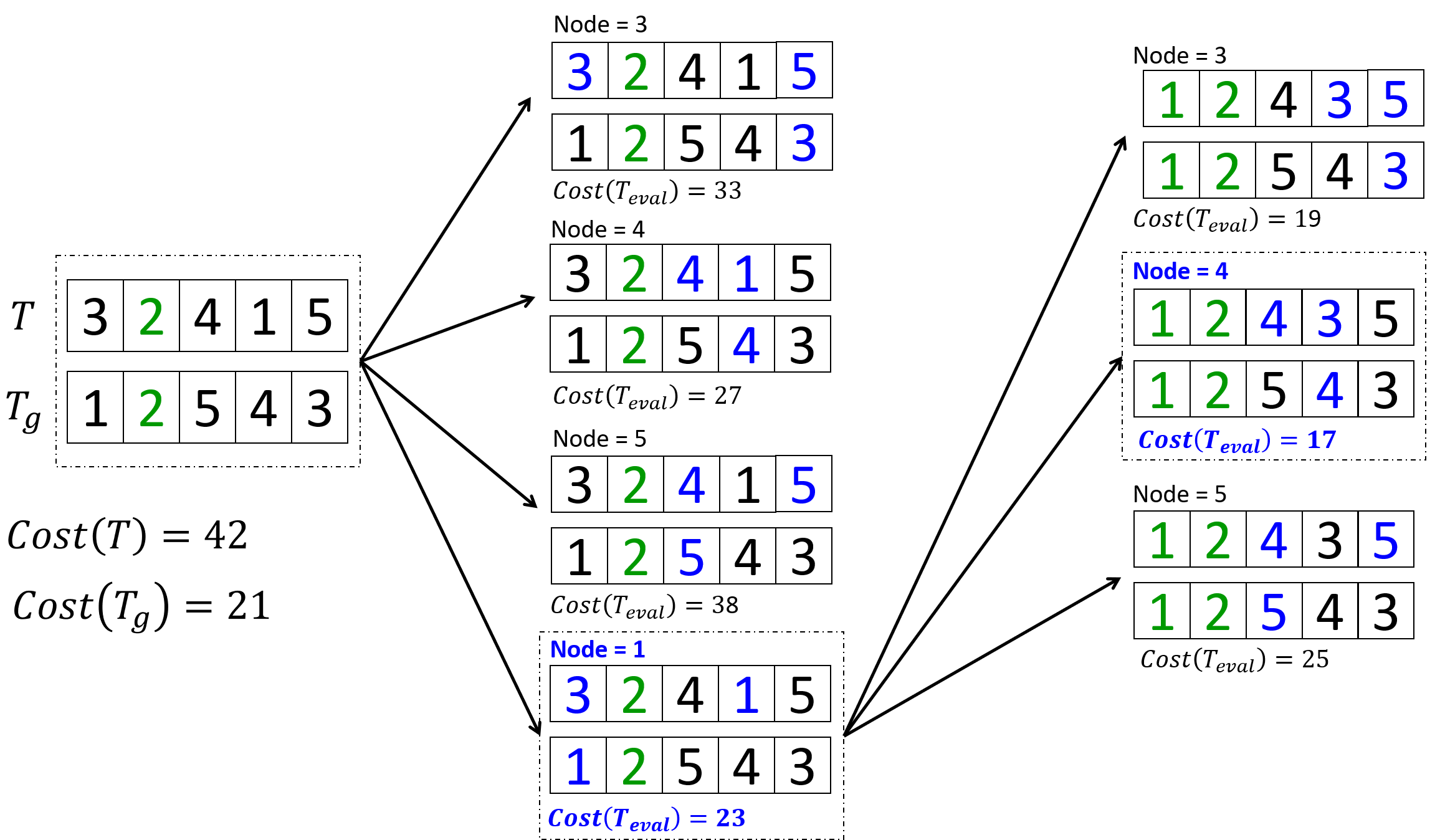} 
    		\caption{The neighborhood evaluation in proposed path relinking. The process to get the best swap position is also described in \Cref{algorithm:get-swap-position-pr}. As also shown in \Cref{algorithm:iteratively-neighborhood-eval} line 7, the intermediate solution $T_{eval}$ will continue to split processes only when $\text{Cost}(T_{eval}) \leq \text{Cost}(T)$}
    		\label{img:pr-neighbors-evaluation}
    	\end{center}
    \end{figure}

    As depicted in \Cref{algorithm:iteratively-neighborhood-eval}, line 6, these neighborhood searches involve swapping the customer nodes towards the guiding solution. The aim is to identify solutions that closely resemble the initial solutions but may potentially yield better objective function values. Moreover, the search in path relinking will automatically terminate whenever the intermediate solution is better than the guiding solution, $\text{Cost}(T_{eval}) \leq \text{Cost}(T_g)$. In our proposed path relinking, we implement the mechanism of truncated path relinking. Therefore, as illustrated in \Cref{algorithm:iteratively-neighborhood-eval}, the number of search moves, denoted as $N_{pr}$, can be assumed to be less than or equal to the size of $L_{pr}$.

    \begin{algorithm}
        \footnotesize
        \caption{Get best swap position for transforming $T$ toward $T_g$.}\label{algorithm:get-swap-position-pr}
        \hspace*{\algorithmicindent} \textbf{input:} current initial solution $T$, guiding solution $T_g$\\
        \hspace*{\algorithmicindent} \qquad \quad list of restricted neighborhood $L_{pr}$, tabu list $L_{tabu}$
        \begin{algorithmic}[1]
            \Procedure{GetPositionSwap}{$T$, $T_g$, $L_{pr}$, $L_{tabu}$}
            \State $c_{swap} \gets -1, p_i \gets -1, p_j \gets -1, f_{best} \gets \infty$ \Comment{initialization}
            \For{$\text{node} \in L_{pr}$}
                \If{$\text{node} \notin L_{tabu}$}
                    \State $pos_i \gets \Call{GetPositionNode}{T,\text{node}}$ \Comment{position of node in $T$}
                    \State $pos_g \gets \Call{GetPositionNode}{T_g,\text{node}}$ \Comment{position of node in $T_g$}
                    \State $\varepsilon \gets \Call{PossibilitySwap}{p_i,p_g}$
                    \If{$\top\varepsilon$}
                        \If{$\Call{CostSwapOn}{pos_i,pos_g} < f_{best}$}
                            \State $f_{best} \gets \Call{CostSwapOn}{pos_i,pos_g}$
                            \State $c_{swap} \gets \text{node}$
                            \State $p_i \gets pos_i$
                            \State $p_j \gets pos_g$
                        \EndIf
                    \EndIf
                \EndIf
            \EndFor
            \State \Return $c_{swap}$, $p_i$, $p_j$
            \EndProcedure
        \end{algorithmic}
    \end{algorithm}
    
\newpage

\begin{landscape}

\subsection{Detailed Test Results on Very Large-Scale Instances}
\label{sec:appendix-result-2}

    \begin{table}[htbp]
        \begin{center}
            \caption{Detailed comparison of solution quality on $\mathbb{B}$ instances \citep{ARNOLD201932}.}
            \label{table:b-detailed-result}
            \vspace*{0.2cm}
            \setlength\tabcolsep{2.2pt}
            \scalebox{0.66}{
                \begin{tabular}{ll r rrr r rrr r rrr r rrr r r}
                    \toprule
                        \multirow{2}{*}{\textbf{Instance}} 
                        && \multicolumn{3}{c}{\textbf{MS}} 
                        && \multicolumn{3}{c}{\textbf{Guided-MS}} 
                        && \multicolumn{3}{c}{\textbf{HGS}} 
                        && \multicolumn{3}{c}{\textbf{HGS$^{*}$}} 
                        && \multirow{2}{*}{\textbf{BKS}} \\
                        
                        \cmidrule(lr){3-5} \cmidrule(lr){7-9} \cmidrule(lr){11-13} \cmidrule(lr){15-17}
                        && \textbf{Avg (Gap)} & \textbf{Best (Gap)} & \textbf{Time} 
                        && \textbf{Avg (Gap)} & \textbf{Best (Gap)} & \textbf{Time} 
                        && \textbf{Avg (Gap)} & \textbf{Best (Gap)} & \textbf{Time} 
                        && \textbf{Avg (Gap)} & \textbf{Best (Gap)} & \textbf{Time} 
                        && \\
                    \midrule
                        Leuven1 && 195946.8 (1.607) & 195946 (1.606) & 900 && 195653.2 (1.455) & 195647 (1.451) & 900  && 195619 (1.437) & 195302 (1.273) & 7200 && 195438 (1.343) & 195101 (1.168) & 7200 && 192848 \\
                        Leuven2 && 120068 (7.790) & 120068 (7.790) & 1200 && 119533 (7.309) & 119533 (7.309) & 1200  && 115236.6 (3.452) & 114933 (3.180) & 9600 && 115074 (3.306) & 114814 (3.073) & 9600  && 111391 \\
                        Antwerp1 && 489454.8 (2.552) & 489449 (2.550) & 1800 && 488045 (2.256) & 488027 (2.252) & 1800  && 486699.8 (1.974) & 486146 (1.858) & 14400  && 486064.6 (1.841) & 485691 (1.763) & 14400  && 477277 \\
                        Antwerp2 && 312221.4 (7.164) & 312194 (7.154) & 2100 && 310386.8 (6.534) & 310368 (6.528) & 2100  && 303638.8 (4.218) & 303047 (4.015) & 16800  && 303267.6 (4.090) & 302773 (3.921) & 16800  && 291350 \\
                        Ghent1 && 481775.6 (2.608) & 481744 (2.601) & 3000 && 479334.2 (2.088) & 479306 (2.082) & 3000  && 483964.4 (3.074) & 483738 (3.026) & 24000 && 483574 (2.991) & 483444 (2.963) & 24000  && 469531 \\
                        Ghent2 && 280948.2 (9.001) & 280915 (8.988) & 3300 && 279330 (8.373) & 279329 (8.373) & 3300  && 271089.4 (5.176) & 270704 (5.027) & 26400 && 270628 (4.997) & 270429 (4.920) & 26400  && 257748 \\
                        Brussels1 && 524784.6 (4.597) & 524216 (4.484) & 4500 && 521452 (3.933) & 521371 (3.917) & 4500  && 521733.6 (3.989) & 521426 (3.928) & 36000 && 521273.6 (3.898) & 520896 (3.822) & 36000  && 501719 \\
                        Brussels2 && 377679.4 (9.324) & 377620 (9.307) & 4800 && 375933.2 (8.819) & 375830 (8.789) & 4800  && 365087.2 (5.679) & 364610 (5.541) & 38400  && 364866.4 (5.615) & 364165 (5.412) & 38400  && 345468 \\
                        Flanders1 && 7462720 (3.075) & 7458480 (3.016) & 6000 && 7425718 (2.563) & 7422360 (2.517) & 6000  && 7478140 (3.288) & 7472490 (3.210) & 48000  && 7472976 (3.216) & 7469230 (3.164) & 48000  && 7240118 \\
                        Flanders2 && 4743088 (8.457) & 4742170 (8.436) & 9000 && 4730592 (8.171) & 4730150 (8.161) & 9000  && 4712134 (7.749) & 4706340 (7.617) & 72000  && 4701680 (7.510) & 4696840 (7.399) & 72000  && 4373244 \\
                    \midrule
                        Average Gap && 5.617 & & && 5.150 & & && 4.004 & & && 3.881 & & && \\
                        Median Gap && 5.881 & & && 5.234 & & && 3.721 & & && 3.602 & & && \\
                    \bottomrule
                \end{tabular}
            }
            \vspace{0.4cm}
            \begin{minipage}{0.98\textwidth}
                {\small \footnotesize $^*$ algorithm with the proposed feature-based guidance.}
            \end{minipage}
        \end{center}
    \end{table}

\end{landscape}

\newpage
\bibliography{ref.bib}

\end{document}